%% file: main.tex
\documentclass[dvipsnames]{article} %
\usepackage{colm2024_conference}

\usepackage{booktabs}
\usepackage{enumitem}
\usepackage{wrapfig}


\usepackage{algorithm}
\usepackage{algpseudocode}

\usepackage{graphicx}
\usepackage[misc]{ifsym}
\usepackage{mathtools}   

\usepackage{microtype}
\usepackage{colortbl}
\usepackage[utf8]{inputenc}
\usepackage[T1]{fontenc}
\definecolor{lightgray}{rgb}{0.9,0.9,0.9}
\usepackage{caption}
\usepackage{subcaption}
\usepackage{setspace}
\usepackage{url}
\usepackage{multirow}
\usepackage{tabularx}
\usepackage{blindtext}
\usepackage{pgfplots}
\pgfplotsset{compat=1.18} 
\usepackage{tikz}
\usetikzlibrary{er,positioning,bayesnet}
\usepackage{makecell}
\usepackage{tipa}
\usepackage{siunitx}
\usepackage{nicefrac}
\usepackage{listings}
\usepackage[raster,skins, most]{tcolorbox} %
\usepackage{xltabular}
\usepackage{adjustbox}
\usepackage{xurl}
\usepackage{rotating}
\usepackage[normalem]{ulem}
\usepackage{xspace}

\usepackage{fontawesome}
\usepackage{pifont}
\usepackage{fancyvrb}
\usepackage{color}
\definecolor{green_first}{RGB}{168, 209, 176}   
\definecolor{green_second}{RGB}{200, 235, 200}  
\definecolor{green_third}{RGB}{235, 255, 235}   
\definecolor{light_green_table}{RGB}{220, 255, 220}  
\definecolor{light_purple_table}{RGB}{235, 225, 255}  
\definecolor{light_green}{rgb}{0.569, 0.800, 0.459}
\definecolor{blue_dist}{rgb}{0.192,0.443,0.651}
\definecolor{dark_purple_table}{RGB}{103, 74, 241}  
\definecolor{purple_cite}{RGB}{1, 2, 129}  

\definecolor{purple_dist}{RGB}{125,94,237}

\definecolor{orange_dist}{rgb}{0.812,0.545,0.239}
\definecolor{yellow_dist}{rgb}{0.918,0.804,0.463}
\definecolor{website}{rgb}{0.9333333333333333, 0.10980392156862745, 0.592156862745098} 
\definecolor{pm_rowcolor}{rgb}{0.85, 0.90, 0.84} 
\definecolor{medium_gray}{RGB}{150, 150, 150}  
\definecolor{medium_purple}{RGB}{150, 120, 200}  
\newtcolorbox{promptbox}[2][Prompt]{
    colback=black!5!white,        
    arc=5pt,                      
    boxrule=0.5pt,                
    fonttitle=\bfseries,          
    title=#1,                     
    before upper={\small},        
    fontupper=\fontfamily{ptm}\selectfont, 
    colframe=#2,                  
    left=3pt,                     
    right=3pt,                    
    top=3pt,                      
    bottom=3pt,                   
    boxsep=3pt,                   
    toptitle=1pt,                 
    bottomtitle=1pt,              
    lefttitle=1pt,                
    righttitle=1pt,               
}

\usepackage{xcolor}      
\newcommand\inb[1]{\colorbox{gray!20}{\lstinline|#1|}}

\useunder{\uline}{\ul}{}

\input{math_commands.tex}

\usepackage{makecell}
\usetikzlibrary{tikzmark}
\makeatletter
\newcommand*\myfontsize{%
  \@setfontsize\myfontsize{7}{8}%
}
\makeatother

\definecolor{uclablue}{RGB}{159, 195, 224}

\definecolor{uclagold}{RGB}{255, 240, 180}

\definecolor{aliceblue}{RGB}{255, 238, 241}

\definecolor{cadmiumgreen}{rgb}{0.0, 0.42, 0.24}

\definecolor{myred}{rgb}{0.7, 0.3, 0.0}
\definecolor{myblue}{rgb}{0.2, 0.3, 0.6}
\definecolor{babygreen}{rgb}{0.85, 0.97, 0.85}

\definecolor{purple1}{RGB}{126, 107, 196}
\definecolor{purple2}{RGB}{199, 158, 207}
\definecolor{purple3}{RGB}{214, 200, 255}
\definecolor{purple4}{RGB}{254, 240, 255}

\definecolor{deepblue}{RGB}{48, 58, 82}

\newcommand{\symboletongyi}{\raisebox{0pt}{~\includegraphics[scale=0.012]{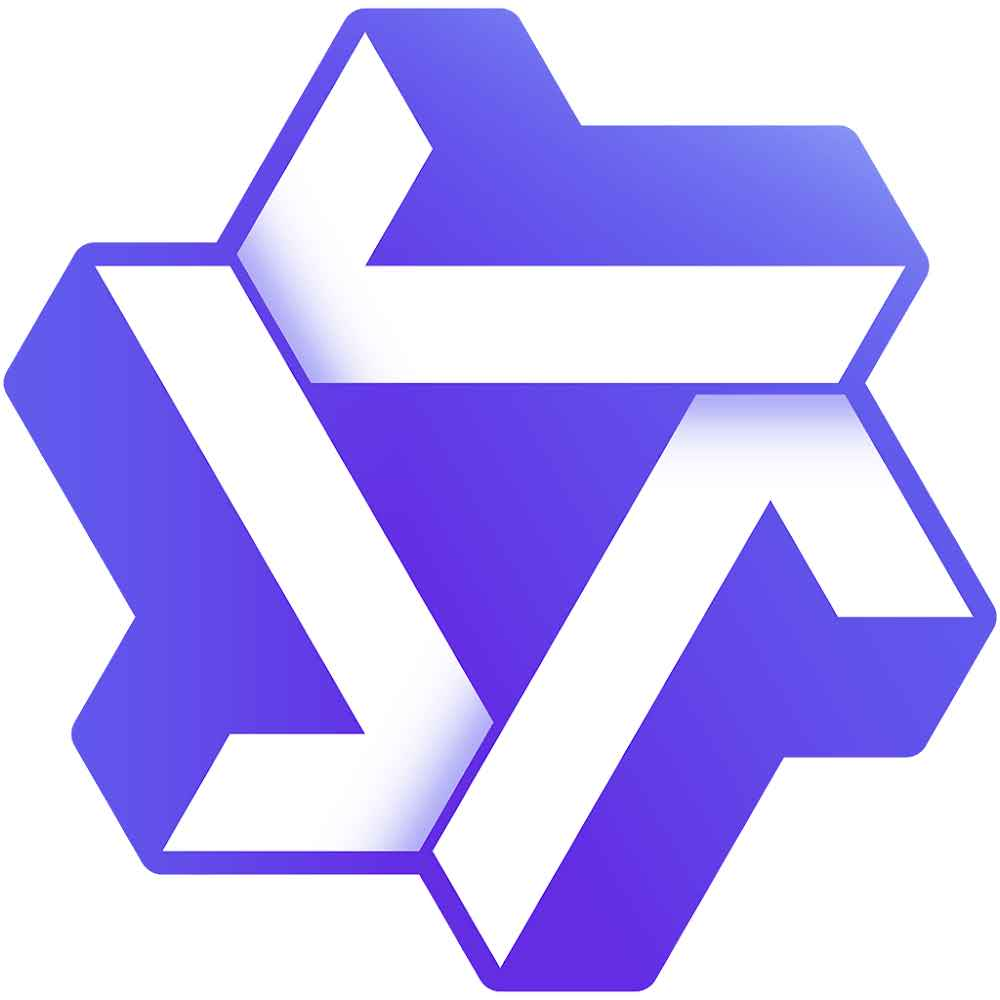}}~}

\definecolor{deepPurple}{HTML}{330066}

\definecolor{uclablue_old}{rgb}{0.15, 0.45, 0.68}
\hypersetup{
    breaklinks,
    citecolor=uclablue_old,
    colorlinks=true,
}

\newtcolorbox{mybox}[2][]
  {colback = black!5!white, colframe = black!75!black, fonttitle = \bfseries,
    colbacktitle = black!100!black, enhanced, before upper={\fontsize{8}{11}\obeyspaces\obeylines\selectfont}, fontupper=\selectfont,
    attach boxed title to top left={yshift=-2.2mm,xshift=4mm},
    title=#2,#1}

\title{%
\begin{tabular}[t]{l}
  \parbox[t]{0.85\textwidth}{\centering
    Reward Modeling from Natural Language Human Feedback
  }
\end{tabular}
}

\author{
Zongqi Wang,
Rui Wang,
Yuchuan Wu,
Yiyao Yu,
Pinyi Zhang,
Shaoning Sun, \\
Yujiu Yang,
Yongbin Li$^{ (\textrm{\Letter})}$
  \\[1em]
  {\fontsize{10pt}{11pt}\selectfont Qwen-Character Team\symboletongyi, Alibaba Group}\\
}

\begin{document}

\maketitle


\input{content/abstract.tex}

\input{content/intro.tex}
\input{content/intro2.tex}

\input{content/method.tex}
\input{content/experiments.tex}
\input{content/related_work.tex}
\input{content/conclusion.tex}


\clearpage
\bibliography{biblio}
\bibliographystyle{colm2024_conference}

\clearpage
\input{content/appendix.tex}

\end{document}

%% file: math_commands.tex
\usepackage{amsmath,amsfonts,bm}

\def\eqref#1{equation~\ref{#1}}

\def\1{\bm{1}}

\DeclareMathAlphabet{\mathsfit}{\encodingdefault}{\sfdefault}{m}{sl}
\SetMathAlphabet{\mathsfit}{bold}{\encodingdefault}{\sfdefault}{bx}{n}



%% file: content/abstract.tex
\begin{abstract}
Reinforcement Learning with Verifiable reward (RLVR) on preference data has become the mainstream approach for training Generative Reward Models (GRMs). Typically in pairwise rewarding tasks, GRMs generate reasoning chains ending with critiques and preference labels, and RLVR then relies on the correctness of the preference labels as the training reward. 
However, in this paper, we demonstrate that such binary classification tasks make GRMs susceptible to guessing correct outcomes without sound critiques. Consequently, these spurious successes introduce substantial noise into the reward signal, thereby impairing the effectiveness of reinforcement learning. 
To address this issue, we propose Reward Modeling from Natural Language Human Feedback (RM-NLHF), which leverages natural language feedback to obtain process reward signals, thereby mitigating the problem of limited solution space inherent in binary tasks. Specifically, we compute the similarity between GRM-generated and human critiques as the training reward, which provides more accurate reward signals than outcome-only supervision. 
Additionally, considering that human critiques are difficult to scale up, we introduce Meta Reward Model (MetaRM) which learns to predict process reward from datasets with human critiques and then generalizes to data without human critiques. 
Experiments on multiple benchmarks demonstrate that our method consistently outperforms state-of-the-art GRMs trained with outcome-only reward, confirming the superiority of integrating natural language over binary human feedback as supervision. 
\end{abstract}

%% file: content/intro.tex
\section{Introduction}

Recently, generative reward models (GRMs) have emerged as a powerful solution to enhance the general capabilities of large language models (LLMs)~\citep{liu2025inference, wang2025helpsteer3, yuan2024self, guo2025reward, huang2025think, xu2025j4r}. 
Unlike traditional scalar reward models that output a single numerical score, GRMs generate natural language rationales with detailed critiques before making judgments, leading to significantly higher robustness, generalizability, and accuracy through explicit reasoning~\citep{mahan2024generative, malik2025rewardbench, tan2024judgebench, liu2024rm, wang2025helpsteer3, ye2024beyond, zhang2025echo, shao2025deepseekmath}. 
Compared to rule-based verifiers that are restricted to verifiable tasks such as mathematical problems~\citep{xie2025logic,shao2024deepseekmath,li2025mtr,yu2025chain,li2025torl} or code execution~\citep{seed2025seed,hui2024qwen2,wang2025epicoder}, GRMs can effectively evaluate a much broader range of tasks including social intelligence~\citep{yu2025sotopia,zhang2025sentient,li2025mimeqa,mao2025mindvote,zhou2025think,zhang2025sentient}, roleplay~\citep{zhou2025personaeval,lu2024large,lu2025rolemrc,qin2025r,nath2025let}, and image generation~\citep{zhang2025generative,wu2025visualquality,zhou2025opening,yang2025self}, etc. 
Typically in a pairwise reward setting, GRMs produce a detailed CoT which mainly serves to derive effective \textbf{critiques}\footnote{In this paper, we use ``process'' to refers to these critiques, instead of full reasoning process.} of two responses, followed by an explicit \textbf{outcome} indicating which response is superior~\citep{yuan2024self, guo2025reward, huang2025think, xu2025j4r}. 

\begin{figure}[t]
    \centering
    \includegraphics[width=1.0\textwidth]{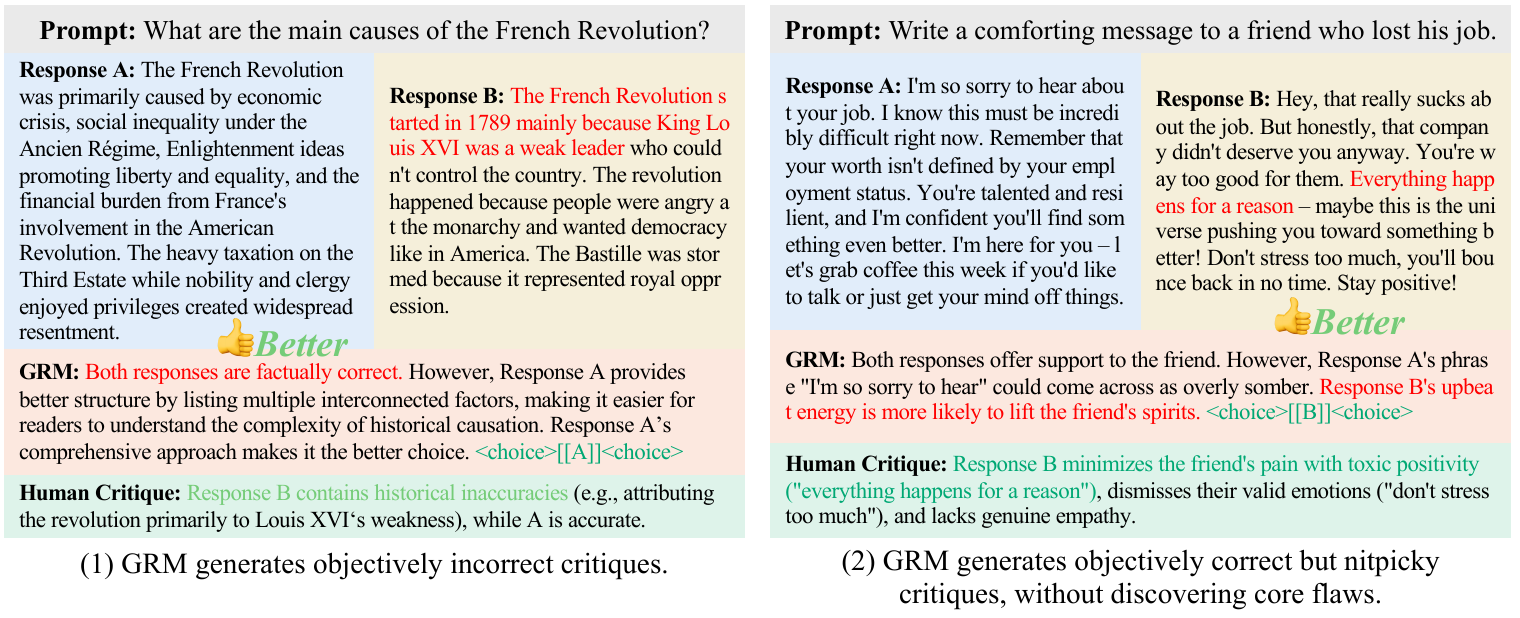}
    \caption{Examples of GRMs achieving correct outcome with flawed critiques.}
    \label{fig:process_outcome_inconsistency}
\end{figure}

The current GRM training in reinforcement learning with verifiable reward (RLVR) typically relies on outcome-only supervision, primarily through binary pairwise preference labels~\citep{wang2025helpsteer3, yuan2024self, guo2025reward}.
\textbf{However, we argue that while outcome reward are effective for mathematical tasks, they are fundamentally inadequate for GRM tasks.} This disparity arises from the nature of their respective \textit{solution spaces}: mathematical tasks typically have large solution spaces where a correct outcome strongly implies correct reasoning trajectory. In contrast, GRM tasks operate within a constrained binary solution space with only two possible outcomes (preference for response A or B). Such a binary classification task makes GRMs susceptible to guessing the correct preference label without generating sound critiques, e.g., objectively incorrect or nitpicky (non-essential) critiques (examples in Figure~\ref{fig:process_outcome_inconsistency}).
We illustrate the mismatch in the proportions of process and outcome supervision between mathematical tasks and GRM tasks in Figure~\ref{fig:motivation_process_outcome_inconsistency}. 
Sometimes, mathematical tasks also exhibit small solution spaces (e.g., multiple-choice questions or true/false problems), and practitioners typically address this by either removing such data~\citep{team2025kimi, cui2025process} or reformulating them as fill-in-the-blank questions to expand the solution space~\citep{albalak2025big}. However, in GRM tasks, the pairwise comparison formulation is inherently essential and cannot be abandoned or restructured. 
\textbf{As a result, given the extremely restricted solution space of binary tasks, these spurious successes introduce substantial noise into the reward signal when using RL to train GRMs, thereby impairing the learning effectiveness.}
Specifically, the outcome-process misalignment resulting from outcome-only supervision might lead the model to exploit spurious correlations rather than developing genuine reasoning capabilities, causing the policy to converge toward producing erroneous critiques and thereby limiting the generalization and advancement potential of GRMs' training. 

In this paper, we first empirically validate the outcome-process inconsistency issue through comprehensive analysis of mainstream GRMs, revealing that a substantial portion (20-30\%) of correct preference predictions are accompanied by flawed critiques, even in state-of-the-art models (Section~\ref{sec:grm_outcome_process_inconsistency}). We then explore various approaches for providing accurate process reward and demonstrate that leveraging the similarity of core critiques between human critiques and GRM-generated critiques as process reward achieves the best performance (Section~\ref{sec:exploring_process_reward_design_for_grm_training}). 

Motivated by these observations, we propose incorporating the similarity between \textbf{\textit{human critiques}} and GRM-generated critiques as an additional process reward term alongside the outcome reward during GRM training. This approach yields significant improvements in both critique quality and outcome accuracy (Section~\ref{sec:human_critique_process_reward_integration}). 
However, human critique annotation faces critical \textbf{\textit{scalability}} challenges due to high costs and the difficulty of obtaining high-quality annotations~\citep{wang2025helpsteer3}. In practice, existing data resources rarely include human critiques, with most datasets containing only outcome labels~\citep{lai2024step,han2024wildguard,wang2024helpsteer,liu2024skywork}. To address this limitation, we propose using \textbf{\textit{MetaRM}}, an auxiliary meta reward model that learns to predict process reward from training data with human critiques and generalizes to unlabeled data with only outcome labels (Section~\ref{sec:metarm_architecture}). 
Additionally, to handle distribution shift as the policy evolves during RL training, we introduce an \textbf{\textit{online MetaRM}} framework that continuously updates the MetaRM alongside GRM updates, enabling more effective process supervision at scale (Section~\ref{sec:online_metarm_training}). Our experiments demonstrate that online MetaRM-based training achieves comparable performance to full human critique supervision while significantly reducing annotation requirements (Section~\ref{sec:metarm_results}).

Our contributions can be summarized as follows:

\begin{itemize}
    \item We empirically demonstrate substantial outcome-process inconsistency in current GRMs, where 20-30\% of correct preference predictions are accompanied by flawed reasoning critiques across multiple state-of-the-art models, motivating the need for explicit process supervision (Section~\ref{sec:grm_outcome_process_inconsistency}).
    
    \item We show that the similarity of core critiques between human critiques and GRM-generated critiques serves as an accurate proxy for process reward. Furthermore, by incorporating such process reward during GRM training, we observe significant improvements over outcome-only supervision (Section~\ref{sec:exploring_process_reward_design_for_grm_training}, Section~\ref{sec:human_critique_process_reward_integration}).
    
    \item To address the challenges of expensive annotation for human critiques, we propose an online MetaRM framework that learns to predict process reward from limited human critiques and generalizes to unlabeled data concurrently with GRM training (Section~\ref{sec:method_metarm}). Through extensive experiments, we demonstrate the remarkable performance of our framework (Section~\ref{sec:experiments}).

    \item We explain four reward modeling mechanisms from \textbf{theoretical perspective}, clarifying the evolution of reward modeling and the necessity of our method, then identify the theory-practice gaps remain (Appendix~\ref{appendix:theory}). 
\end{itemize}

%% file: content/intro2.tex
\section{Problem Formulation and Motivation}
\label{sec:problem_formulation_and_motivation}

In this section, we first formulate the pairwise rewarding tasks and provide a review of traditional RL with outcome reward (Section~\ref{sec:formulation_grm_rl}). 
Next, we empirically demonstrate the inconsistency between the current GRM's outcome and process (Section~\ref{sec:grm_outcome_process_inconsistency}). 
Finally, we explore various approaches that can serve as proxies for process reward (Section~\ref{sec:exploring_process_reward_design_for_grm_training}). 

\subsection{Formulation of Pairwise Rewarding Task and GRM Training with Outcome Reward}
\label{sec:formulation_grm_rl}

Pairwise rewarding tasks are designed to evaluate the comparative quality of two response candidates. 
A sample consists of a query $q$ and two candidate responses, $y_A$ and $y_B$, along with a preference label $l \in \{A, B\}$ indicating which response is superior. 
The objective of GRM \( \pi_\theta \) is to generate reasoning with critiques for two candidates and predict a preference label \( \hat{l} \) matching the ground truth. 

In RL with outcome reward under the GRPO framework, the model receives a binary reward $R_{\text{outcome}} \in \{0, 1\}$ based on whether its predicted preference label matches the ground truth. 
Training proceeds by generating $N_{\text{rollout}}$ candidate responses for the same query, computing the average reward $\bar{R}$ and standard deviation $\sigma$ within the group, and then normalizing reward to obtain advantages:

\begin{equation}
\bar{R} = \frac{1}{N_\text{rollout}}\sum_{i=1}^{N_\text{rollout}}{{R_{\text{outcome}}}_i}, \quad
\sigma=\frac{1}{N_\text{rollout}}\sqrt{\sum_{i=1}^{N_\text{rollout}}{({R_{\text{outcome}}}_i-\bar{R})^2}},
\end{equation}
\begin{equation}
\hat{A}_{i} = \frac{{R_{\text{outcome}}}_{i} - \bar{R}}{\sigma}.
\end{equation}
The advantage $\hat{A}_i$ is assigned uniformly to all tokens in a response, and the policy gradient update is computed using clipping and KL regularization to stabilize optimization:
\begin{equation}
\mathcal{J}(\theta)=\mathbb{E}\left[\frac{1}{G}\sum_{i=1}^{G}\frac{1}{|\hat{y}_i|}\sum_{t=1}^{|\hat{y}_i|}
\left(\min\left(r_{i,t}\hat{A}_{i},\text{clip}\left(r_{i,t}, 1-\varepsilon, 1+\varepsilon \right)\hat{A}_{i} \right) -\beta D_{\text{kL}} \right)\right].
\label{eq:grpo_loss}
\end{equation}

\subsection{Outcome–Process Inconsistency}
\label{sec:grm_outcome_process_inconsistency}

To validate the outcome-process inconsistency, we conduct experiments on two distinct tasks: mathematical reasoning on MATH-500~\citep{hendrycks2measuring} and pairwise rewarding on validation set of HelpSteer3~\citep{wang2025helpsteer3}. 
For mathematical reasoning, we use gemini-2.5-pro to compare model-generated solutions against ground-truth solutions, assessing both process validity and answer correctness. 
For pairwise rewarding, we evaluate outcome accuracy using a rule-based verifier and process accuracy by computing the similarity between model-generated critiques and human critiques using gemini-2.5-pro (whose accuracy we validate in Section~\ref{sec:exploring_process_reward_design_for_grm_training}). 
%

\begin{figure}[h]
    \centering
    \begin{subfigure}[t]{0.385\textwidth}
        \centering
        \includegraphics[width=\textwidth]{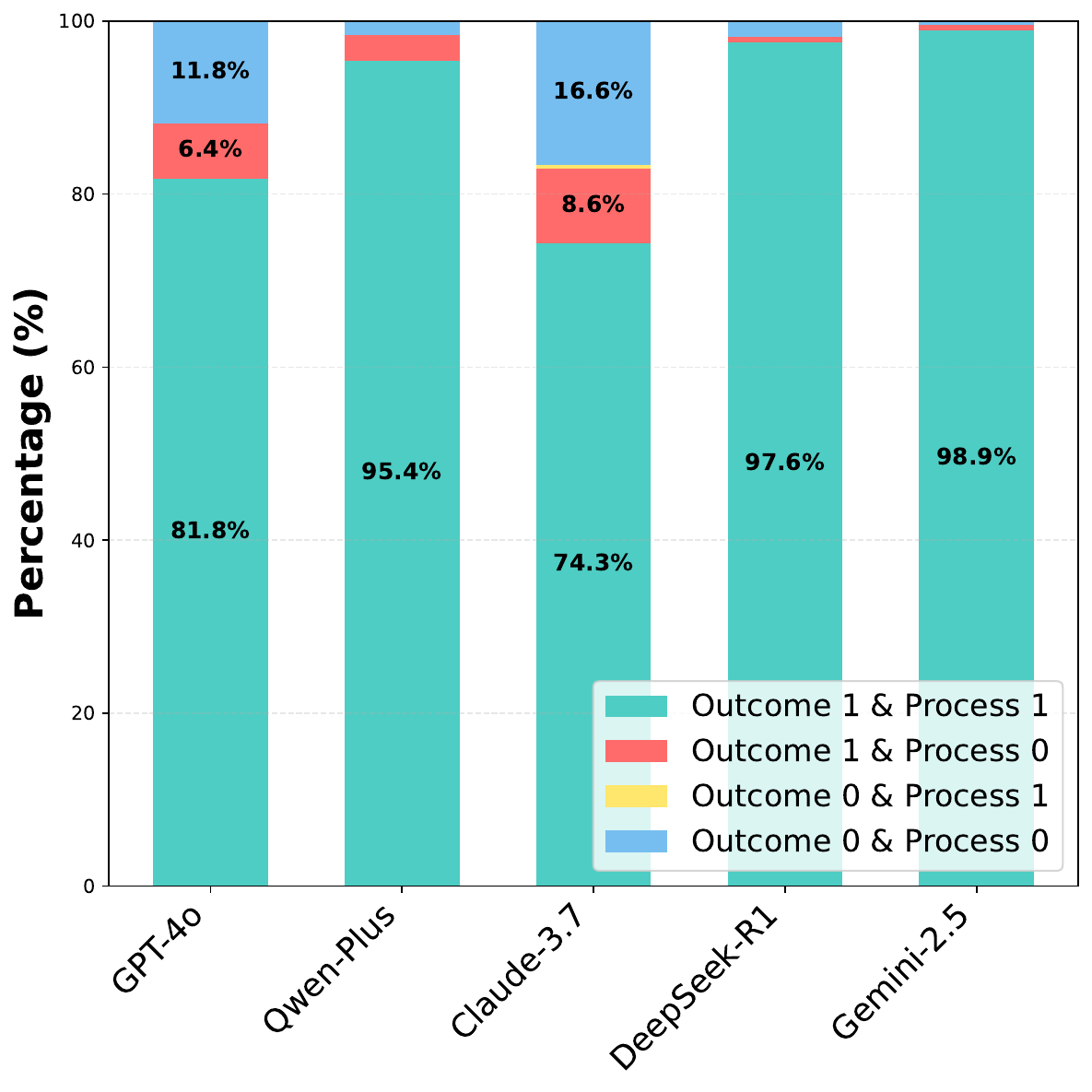}
        \caption{Math tasks.}
        \label{fig:inconsistency_math}
    \end{subfigure}
    \hspace{1.0em}
    \begin{subfigure}[t]{0.48\textwidth}
        \centering
        \includegraphics[width=\textwidth]{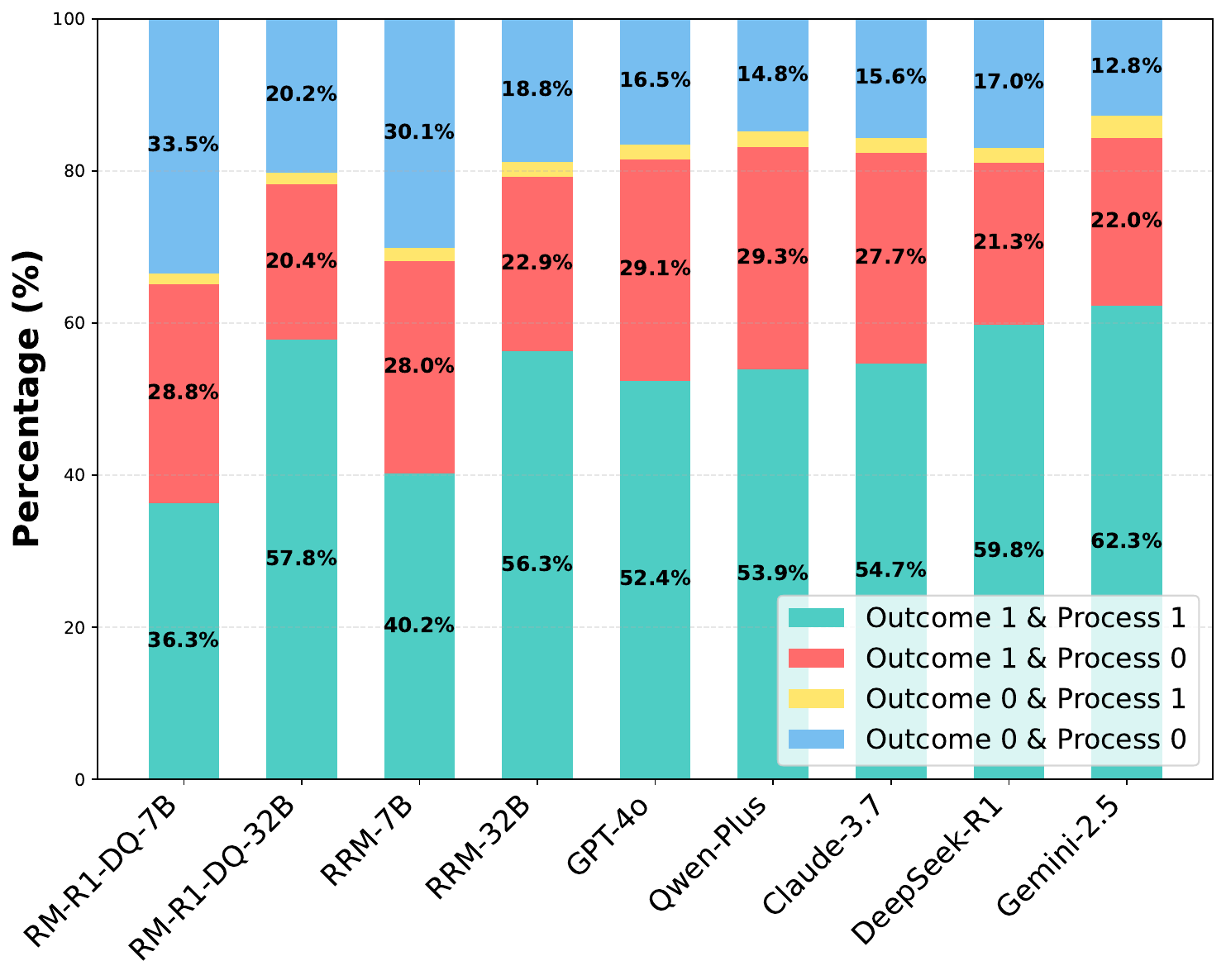}
        \caption{Pairwise rewarding tasks.}
        \label{fig:inconsistency_preference}
    \end{subfigure}
    \caption{Inconsistency between the correctness of outcome and process across various models and tasks.}
    \label{fig:motivation_process_outcome_inconsistency}
\end{figure}

We present the results in Figure~\ref{fig:motivation_process_outcome_inconsistency}, which reveal striking patterns that differ between the two task types:

\textbf{Mathematical Reasoning Tasks.} The probability of outcome-process inconsistency is extremely low across all models. This is because mathematical reasoning has a large solution space, making it difficult to arrive at correct answers through invalid reasoning. 

\textbf{Pairwise Rewarding Tasks.} Models frequently produce correct preference labels $l$ but generate flawed critiques $\hat{c}$. For example, RM-R1-DeepSeek-Distilled-Qwen-7B shows this probability at 44.24\%, while strong proprietary models like gemini-2.5-pro and claude-3.7-sonnet exhibit 26.10\% and 33.62\%, respectively. 
Additionally, low $P(\text{process}=1|\text{outcome}=0)$ suggests that incorrect outcomes almost always correspond to incorrect critiques, which will also motivate our method design in Section~\ref{sec:human_critique_process_reward_integration}. 


\begin{tcolorbox}[colframe=dark_purple_table, colback=white, coltitle=purple_cite]
\textcolor{dark_purple_table}{\textbf{Takeaway from Section~\ref{sec:grm_outcome_process_inconsistency}}}\\
GRMs frequently produce correct outcomes despite generating invalid critiques. This high outcome-process inconsistency suggests that using outcome reward as training signals for GRMs will introduce noise, necessitating explicit process supervision. 
Notably, an incorrect outcome almost invariably indicates flawed critiques. 
\end{tcolorbox}

\subsection{Exploring Proxy for Process Reward}
\label{sec:exploring_process_reward_design_for_grm_training}

In this section, we explore how to design accurate proxy for process reward. 
To this end, we first carefully curate and annotate a small set of 49 samples from HelpSteer3. This dataset comprises queries $q$, response pairs $y_a$ and $y_b$, GRM-generated critiques $\hat{c}$, human critiques $h$ originally provided in HelpSteer3, and a label $z \in \{0, 1\}$ indicating whether the critique $\hat{c}$ is correct. 

We investigate three candidate approaches for obtaining process reward: (1) \textbf{LLM-as-a-Meta-Judge}, which employs an external LLM to directly evaluate the correctness of $\hat{c}$; (2) \textbf{Similarity w/ All HC}, which uses an external LLM to extract all critiques mentioned in $h$ and $\hat{c}$, then measures their similarity through three variants: F1, Recall, and Precision; (3) \textbf{Similarity w/ Core HC}, which employs an external LLM to identify and extract only the core arguments (excluding nitpicky critiques) in $h$ and $\hat{c}$, then computes their similarity. For brevity, we defer the detailed experimental setup to Appendix~\ref{appendix:exploring_process_reward_design_for_grm_training}. 

\begin{table}[h]
\centering
\caption{Accuracy (\%) of different methods for process reward. We adopt three variants (F1, Recall, and Precision) to calculate similarity between human and GRM-generated critiques.}
\label{tab:performance_process_reward}
\resizebox{0.9\textwidth}{!}{
\begin{tabular}{l|c||c||ccc|ccc}
\toprule
    \multirow{3}{*}{\textbf{Model}} & \multirow{2}{*}{\makecell{\textbf{Cost} \\ \textbf{(\$/1M Token)}}} & \textbf{LLM-as-a-Meta-Judge} & \multicolumn{3}{c|}{\textbf{Similarity w/ All HC}} & \multicolumn{3}{c}{\textbf{Similarity w/ Core HC}} \\
    \cmidrule{3-9}
    & & \textbf{-} & \textbf{F1} & \textbf{Recall} & \textbf{Precision} & \textbf{F1} & \textbf{Recall} & \textbf{Precision} \\
    \midrule
    gemini-2.5-pro & 15.00 & 0.7347 & 0.8571 & \textbf{0.9388} & 0.7959 & \textit{\underline{0.9184}} & \textit{\underline{0.9184}} & 0.8980 \\
    gpt-5-mini & 2.00 & 0.4898 & 0.7755 & \textit{\underline{0.8163}} & 0.7347 & \textbf{0.8571} & \textbf{0.8571} & \textit{\underline{0.8163}} \\
    gpt-4o-mini & 0.60 & 0.2653 & \textit{\underline{0.7551}} & \textit{\underline{0.7551}} & \textbf{0.7755} & 0.7347 & 0.7347 & \textit{\underline{0.7551}} \\
\bottomrule
\end{tabular}
}
\end{table}

Table~\ref{tab:performance_process_reward} presents the performance of different process reward design approaches. We observe that the LLM-as-a-Meta-Judge approach is substantially outperformed by Similarity w/ HC-based methods across all models, with performance gaps ranging from 12\% to 59\%. On average, Similarity w/ Core HC achieves better performance than Similarity w/ All HC, indicating that focusing on essential critical points provides more reliable process supervision. While the highest performance (0.9388) is achieved by gemini-2.5-pro with Similarity w/ All HC using Recall, we adopt the more cost-effective gpt-5-mini with Similarity w/ Core HC and F1 metric, which balances accuracy, cost-effectiveness, and training stability (Recall tends to cause reward hacking as discussed in Appendix~\ref{appendix:discussion_reward_hacking}). 

\begin{tcolorbox}[colframe=dark_purple_table, colback=white, coltitle=purple_cite]
\textcolor{dark_purple_table}{\textbf{Takeaway from Section~\ref{sec:exploring_process_reward_design_for_grm_training}}}\\
To address the outcome-process inconsistency, we need an accurate proxy for process reward. 
Our experiments reveal that computing the F1-based similarity between core arguments in human critiques and GRM-generated critiques serves as an effective proxy for process reward. 
\end{tcolorbox}

%% file: content/method.tex
\section{Reward Modeling from Natural Language Human Feedback}
\label{sec:method}

This section is organized as follows. 
First, we directly use the F1-based similarity between human critiques and GRM-generated critiques as process reward to guide GRM training, achieving promising results (Section~\ref{sec:human_critique_process_reward_integration}).
Next, we propose MetaRM to address the scalability challenges of human critique annotation (Section~\ref{sec:metarm_architecture}). 
Finally, to address the issue that distribution shifts in GRM outputs during RL training lead to inaccuracies in MetaRM, we propose an online MetaRM framework (Section~\ref{sec:online_metarm_training}).

\subsection{Natural Language Human Critique as Process Reward}
\label{sec:human_critique_process_reward_integration}

Motivated by the findings in Section~\ref{sec:exploring_process_reward_design_for_grm_training}, we use the F1-based similarity between human critiques and GRM-generated critiques as a process reward signal. In this section, we incorporate this signal into GRM training via a composite reward that combines process- and outcome-level supervision. 

\paragraph{Reward Design.} The process reward is defined as: 
\begin{equation}
R_{\text{process}} = \begin{cases}
1, & \text{if } S(h, \hat{c}) > 0.5,  \\
0, & \text{otherwise,}
\end{cases}
\label{eq:process_reward}
\end{equation}
where $S(h, \hat{c})$ measures the F1-based similarity of core arguments of human critiques $h$ and GRM-generated critiques $\hat{c}$ (we adpot gpt-5-mini as calculator here). 

To incorporate human critiques for process supervision while maintaining outcome correctness, we design a composite reward function that balances both aspects. The final reward is defined as:
\begin{equation}
R = \begin{cases}
-1, & \text{if output format is invalid,} \\
0, & \text{if } \hat{l} \neq l, \\
1 + \lambda \cdot R_{\text{process}}, & \text{if } \hat{l} = l,
\end{cases}
\label{eq:composite_reward}
\end{equation}
where $\lambda \in [0, 1]$ controls the weight of process supervision. We adopt \textbf{outcome regularization}, i.e., we exclude process reward when outcomes are incorrect since they indicate invalid critiques (Section~\ref{sec:grm_outcome_process_inconsistency}). 

\paragraph{Experimental Results.} 
We train the model using GRPO for one epoch on the HelpSteer3 training set. To evaluate the impact of incorporating human critiques as process reward, we compare training with and without human critiques on the HelpSteer3 validation set (Figure~\ref{fig:human_critique_effectiveness}). Detailed experimental settings are provided in Appendix~\ref{appendix:exp_preliminary_details}. 

\begin{figure}[h]
    \centering
    \begin{subfigure}[t]{0.48\textwidth}
        \centering
        \includegraphics[width=\textwidth]{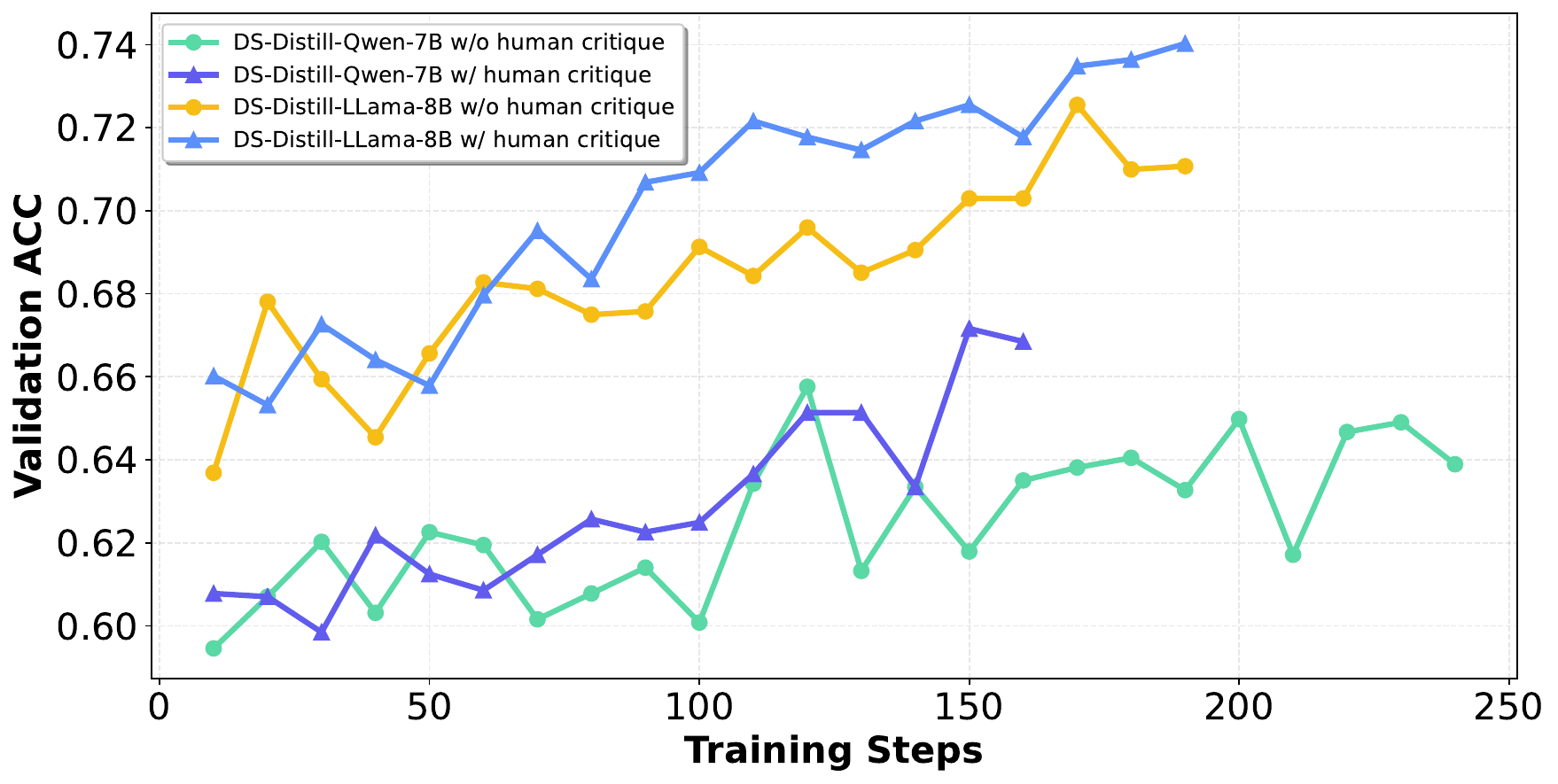}
    \end{subfigure}
    \hfill
    \begin{subfigure}[t]{0.48\textwidth}
        \centering
        \includegraphics[width=\textwidth]{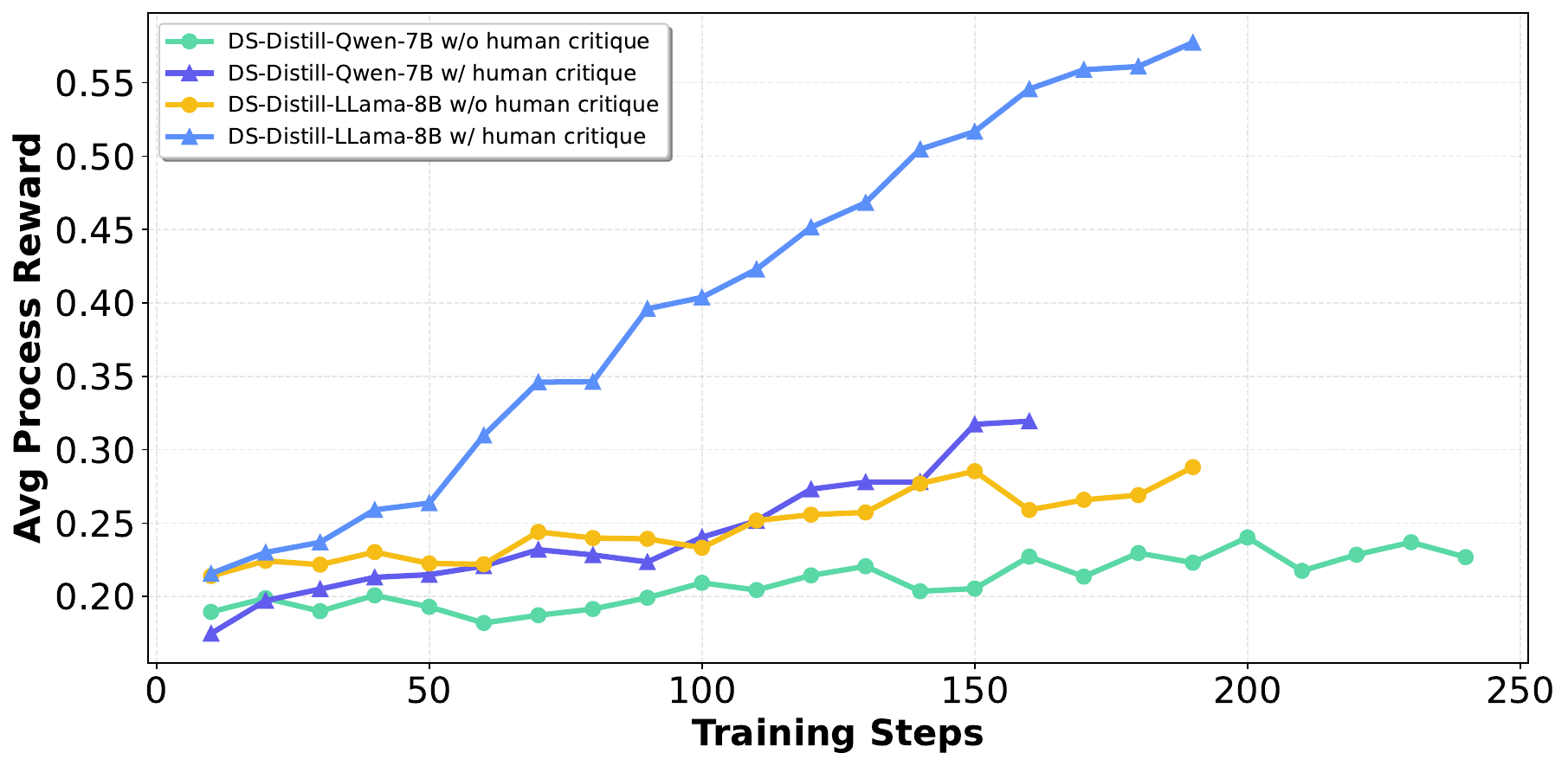}
    \end{subfigure}
    \caption{Comparison between using human critique as process reward and outcome-only reward.}
    \label{fig:human_critique_effectiveness}
\end{figure}

We observe that, for both Deepseek-R1-Distill-Llama-8B and Deepseek-R1-Distill-Qwen-7B, incorporating human critiques as process reward consistently outperforms outcome-only training. Models trained with human critiques begin to show stable improvements after approximately 70 steps and maintain this advantage for the remainder of training. 

\begin{tcolorbox}[colframe=dark_purple_table, colback=white, coltitle=purple_cite]
\textcolor{dark_purple_table}{\textbf{Takeaway from Section~\ref{sec:human_critique_process_reward_integration}}}\\
Incorporating F1-based similarity of core arguments between human and GRM-generated critiques as process reward improves GRMs' performance compared to using outcome-only reward. 
\end{tcolorbox}

\subsection{Online MetaRM as Process Reward}
\label{sec:method_metarm}

\begin{figure}[h]
    \centering
    \includegraphics[width=0.9\textwidth]{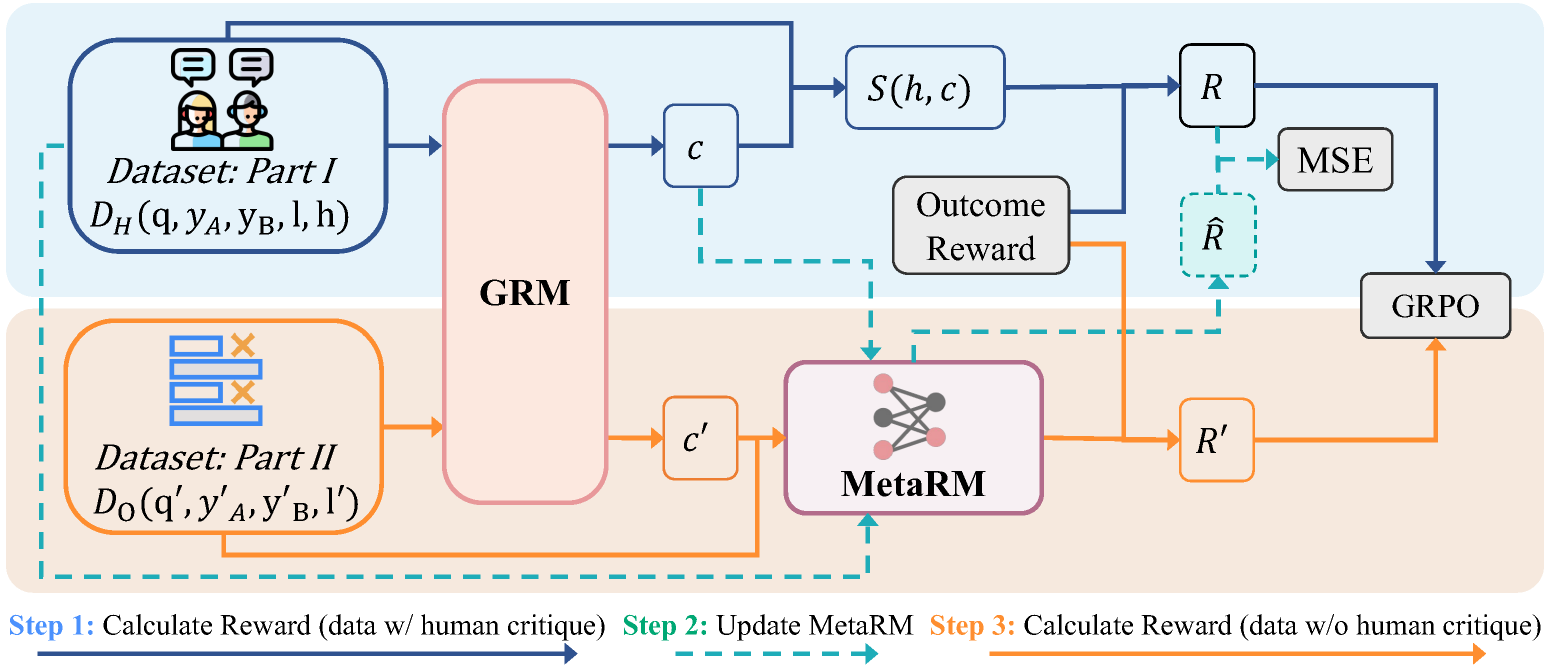}
    \caption{Workflow of online MetaRM framework. The formal algorithm is summarized in Algorithm~\ref{alg:online_metarm}.}
    \label{fig:workflow_online_metarm}
\end{figure}

\subsubsection{Overview}
\label{sec:metarm_overview}

Incorporating human critiques as process reward has been shown to be effective for guiding GRM training. However, collecting such critiques is expensive, which poses significant challenges for scaling this approach to large datasets. 
To address this limitation, we propose \textit{online MetaRM}, a scalable framework that predicts process reward from limited human critique data and generalizes to samples without human critiques. MetaRM leverages a small labeled subset of data, $\mathcal{D}_H$, containing human critiques to train a scalar reward model, which can then provide process supervision for the larger unlabeled dataset, $\mathcal{D}_O$, containing only outcome labels. 

As illustrated in Figure~\ref{fig:workflow_online_metarm}, our framework operates in a three-step process. 
First, for samples in $\mathcal{D}_H$, process reward are computed by measuring the similarity between humans and GRM-generated critiques using Equation~\ref{eq:composite_reward}. 
Second, these computed reward serve as supervision signals to train the MetaRM. 
Third, for samples in $\mathcal{D}_O$, the trained MetaRM predicts process reward, providing supervision where human critiques are unavailable. This design allows the GRM to be trained with complementary reward signals: human-critique-based reward for $\mathcal{D}_H$ and MetaRM-predicted reward for $\mathcal{D}_O$. 

\subsubsection{MetaRM Architecture and Training}
\label{sec:metarm_architecture}

\paragraph{Architecture.}
The MetaRM $M_\phi$ is designed as a regression model that learns to predict reward combining both outcome and process information. Given a query $q$, two candidate responses $y_A$ and $y_B$, and a GRM-generated critique $\hat{c}$, the MetaRM outputs a scalar reward score:
\begin{equation}
\hat{R}_{\text{meta}} = M_\phi(q, y_A, y_B, \hat{c}) \in [0, 1+\lambda],
\label{eq:infer_metarm}
\end{equation}
where $\lambda$ is the weight of process reward defined in Equation~\ref{eq:composite_reward}.

\paragraph{Training Objective.}
For $\mathcal{D}_H$ containing human critiques $h$, the target reward $R_{\text{target}}$ is computed as $R$ defined in Equation~\ref{eq:composite_reward}. The MetaRM is optimized using mean squared error (MSE):
\begin{equation}
\mathcal{L}_{\text{MetaRM}}(\phi) = \mathbb{E}_{(q,y_A,y_B,\hat{c},R_{\text{target}})} \left[ \left( M_\phi(q, y_A, y_B, \hat{c}) - R_{\text{target}} \right)^2 \right].
\label{eq:metarm_loss}
\end{equation}

\begin{figure}[t]
    \centering
    \includegraphics[width=1.0\textwidth]{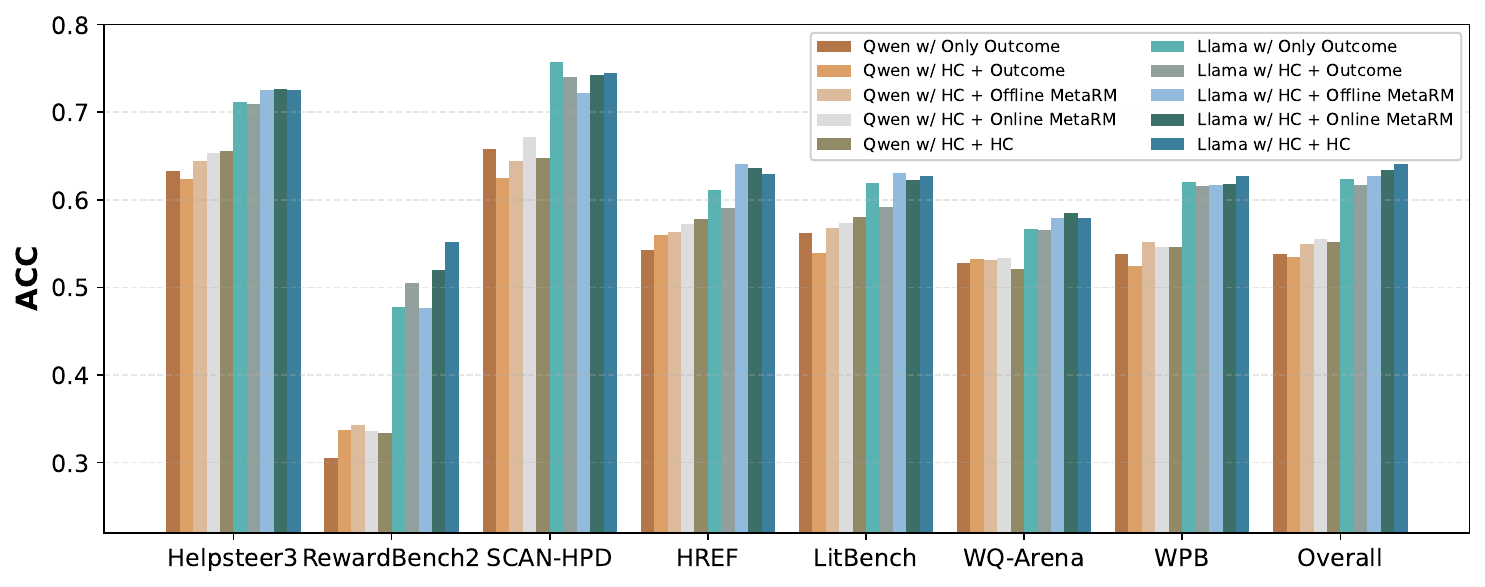}
    \caption{Comparison of offline and online MetaRM variants across multiple benchmarks. }
    \label{fig:metarm_outcome_multi_benchmarks}
\end{figure}

\paragraph{Inference.}
Once trained, the MetaRM can predict process reward for samples without human critiques. For a new sample $(q, y_A, y_B)$ with GRM-generated critique $\hat{c}$, we compute:
\begin{equation}
R' = \begin{cases}
-1, & \text{if output format is invalid,} \\
0, & \text{if } \hat{l} \neq l, \\
1 + \max(0, \hat{R}_{\text{meta}} - 1), & \text{if } \hat{l} = l,
\end{cases}
\label{eq:metarm_reward}
\end{equation}
This design aligns with Equation~\ref{eq:composite_reward} and ensures that correct outcomes receive base reward 1, augmented by the process component predicted by MetaRM, clipped to $[0, \lambda]$. For MetaRM, we also adopt \textbf{outcome regularization}, i.e., no process reward if outcome is incorrect. 

\paragraph{Cold-start Initialization.}
To initialize the MetaRM, we sample $N_{\text{sample}}$ responses from the base model for each sample in $\mathcal{D}_H$. We then compute the reward for these samples according to Equation~\ref{eq:composite_reward} and optimize the model using the objective function in Equation~\ref{eq:metarm_loss} to obtain the initial MetaRM ${M_{\phi}}_{\text{cold}}$.

\subsubsection{Online MetaRM Updating}
\label{sec:online_metarm_training}

Inspired by previous work~\citep{cui2025process, hong2025cooper}, while the offline MetaRM ${M_{\phi}}_{\text{cold}}$ provides a useful initial model, it faces a critical challenge: distribution shift as the policy model evolves during RL training. As the GRM's output distribution of critiques changes over training, the MetaRM trained on initial critiques may become miscalibrated when evaluating later-stage critiques. 
To address this challenge, we propose online updating that continuously adapts the MetaRM alongside policy updates. At each training iteration, we first update the MetaRM using samples from $\mathcal{D}_H$ to calibrate it to the current GRM's output distribution, then use the updated MetaRM to predict process reward for remaining dataset $\mathcal{D}_O$. This MetaRM-first update strategy ensures that the reward model remains aligned with the evolving policy throughout training, enabling continuous effective process supervision. We provide complete algorithm and additional implementation details in Appendix~\ref{appendix:online_metarm}.

\subsubsection{Experimental Evaluation of Online MetaRM}
\label{sec:metarm_results}

Results shown in Figure~\ref{fig:metarm_outcome_multi_benchmarks} (implementation details in Appendix~\ref{appendix:online_metarm}) lead to the following conclusions: 


\textbf{Naive Combination is Catastrophic.} Simply applying process reward to data with human critiques while using outcome-only reward for other data performs even worse than using outcome-only reward for both dataset. We attribute this to conflicting reward signals that undermine reward consistency. 

\textbf{Online MetaRM Outperforms Offline MetaRM.} Online MetaRM consistently outperforms its offline counterpart, highlighting the importance of continuous adaptation to the evolving critique distribution. 

\textbf{Online MetaRM Achieves Performance with Full Human Critique Supervision.} Compared to using human critiques for both datasets, online MetaRM achieves comparable results while significantly reducing annotation costs. Note that this conclusion may not hold for out-of-distribution settings, which we will thoroughly investigate in subsequent experiments. 

Additionally, we include extensive \textbf{ablation studies on MetaRM design} choices in Table~\ref{tab:metarm_ablation} and present the \textbf{accuracy of offline/online MetaRM} during training in Figure~\ref{fig:acc_online_offline_metarm}. 

\begin{tcolorbox}[colframe=dark_purple_table, colback=white, coltitle=purple_cite]
\textcolor{dark_purple_table}{\textbf{Takeaway from Section~\ref{sec:method_metarm}}}\\
We find that naively assigning different reward to different datasets proves catastrophic to performance. In contrast, online MetaRM substantially outperforms offline methods and achieves performance comparable to full human supervision while requiring significantly fewer resources. 
\end{tcolorbox}

%% file: content/experiments.tex
\section{Experiments}
\label{sec:experiments}

\subsection{Experimental Setup}

We evaluate on 6 benchmarks released within one month before the base model's release to avoid data contamination. Our training set contains 164K pairwise preference samples, using GRPO for RL training. We compare RM-NLHF against zero-shot LLM-as-a-Judge, scalar reward models and GRMs trained with same base models (RM-R1, RRM) and  baselines. Details are in Appendix~\ref{appendix:exp_details}.

\subsection{Main Results}

\begin{table}[h]
\centering
\caption{Performance Comparison across Multiple Benchmarks.}
\label{tab:main_results}
\resizebox{\textwidth}{!}{
\begin{tabular}{l|ccccccc|c}
    \toprule
    \textbf{Model} & \textbf{HelpSteer3} & \textbf{Reward Bench V2} & \textbf{SCAN-HPD} & \textbf{HREF} & \textbf{LitBench} & \textbf{WQ\_Arena} & \textbf{WPB} & \textbf{Overall} \\
    \midrule
    \multicolumn{9}{c}{\cellcolor[HTML]{D7D1FF}\textit{Base Models (LLM-as-a-Judge)}} \\
    \midrule
    gpt-5-2025-08-07 & 0.8245 & 0.8344 & 0.8147 & 0.7595 & 0.7617 & 0.6304 & 0.5939 & \cellcolor[HTML]{FEF0CC}0.7456 \\
    o3-2025-04-16 & 0.8315 & 0.8009 & 0.8147 & 0.7058 & 0.7202 & 0.6432 & 0.6234 & \cellcolor[HTML]{FEF0CC}0.7342 \\
    gemini-2.5-pro & 0.8207 & 0.7117 & 0.8288 & 0.6934 & 0.7181 & 0.6529 & 0.6367 & \cellcolor[HTML]{FEF0CC}0.7232 \\
    claude-3-7-sonnet-20250219 & 0.8042 & 0.7714 & 0.7524 & 0.7384 & 0.7631 & 0.6300 & 0.5967 & \cellcolor[HTML]{FEF0CC}0.7223 \\
    deepseek-r1-0528 & 0.7827 & 0.7292 & 0.8227 & 0.6496 & 0.7000 & 0.6348 & 0.6183 & \cellcolor[HTML]{FEF0CC}0.7053 \\
    qwen-plus-latest & 0.8038 & 0.7717 & 0.8048 & 0.6488 & 0.6835 & 0.5961 & 0.6278 & \cellcolor[HTML]{FEF0CC}0.7052 \\
    qwen3-max & 0.8102 & 0.7169 & 0.7865 & 0.6219 & 0.6859 & 0.6062 & 0.6133 & \cellcolor[HTML]{FEF0CC}0.6916 \\
    gpt-4o-latest & 0.7896 & 0.6478 & 0.7572 & 0.5777 & 0.6319 & 0.5799 & 0.5883 & \cellcolor[HTML]{FEF0CC}0.6532 \\
    deepseek-v3 & 0.7522 & 0.5436 & 0.7636 & 0.5678 & 0.6536 & 0.5848 & 0.6122 & \cellcolor[HTML]{FEF0CC}0.6397 \\
    DeepSeek-R1-Distill-Qwen-7B & 0.5999 & 0.3052 & 0.6015 & 0.5487 & 0.5418 & 0.5365 & 0.5161 & \cellcolor[HTML]{FEF0CC}0.5214 \\
    R1-Distill-Llama-8B & 0.6400 & 0.4095 & 0.6592 & 0.5899 & 0.5822 & 0.5650 & 0.5841 & \cellcolor[HTML]{FEF0CC}0.5757 \\
    R1-Distill-Qwen-32B & 0.7376 & 0.6154 & 0.7492 & 0.6967 & 0.6206 & 0.6115 & 0.6389 & \cellcolor[HTML]{FEF0CC}0.6671 \\
    \midrule
    \multicolumn{9}{c}{\cellcolor[HTML]{D7D1FF}\textit{Scalar Reward Models}} \\
    \midrule
    ArmoRM-Llama3-8B-v0.1 & 0.7640 & 0.6897 & 0.6628 & 0.7322 & 0.7157 & 0.5855 & 0.4495 & \cellcolor[HTML]{FEF0CC}0.6571 \\
    URM-LLaMa-3.1-8B & 0.8012 & 0.8004 & 0.7013 & 0.7347 & 0.6302 & 0.5814 & 0.4884 & \cellcolor[HTML]{FEF0CC}0.6768 \\
    Skywork-Reward-Llama-3.1-8B-v0.2 & 0.7950 & 0.7907 & 0.7205 & 0.7347 & 0.6593 & 0.5743 & 0.5322 & \cellcolor[HTML]{FEF0CC}0.6867 \\
    INF-ORM-Llama3.1-70B & 0.8075 & 0.8066 & 0.7684 & 0.7364 & 0.7141 & 0.5978 & 0.5967 & \cellcolor[HTML]{FEF0CC}0.7182 \\
    \midrule
    \multicolumn{9}{c}{\cellcolor[HTML]{D7D1FF}\textit{Specialized Generative Reward Models}} \\
    \midrule
    RM-R1-Qwen-7B & 0.6499 & 0.4679 & 0.6438 & 0.6504 & 0.5750 & 0.5181 & 0.5261 & \cellcolor[HTML]{FEF0CC}0.5759 \\
    RRM-Qwen-7B & 0.6794 & 0.5082 & 0.6645 & 0.6777 & 0.5383 & 0.5421 & 0.5161 & \cellcolor[HTML]{FEF0CC}0.5895 \\
    RRM-Qwen-32B & 0.7942 & 0.7340 & 0.7604 & 0.7273 & 0.6746 & 0.5875 & 0.6283 & \cellcolor[HTML]{FEF0CC}0.7009 \\
    RM-R1-Qwen-32B & 0.7818 & 0.7260 & 0.7795 & 0.7099 & 0.6934 & 0.5988 & 0.6367 & \cellcolor[HTML]{FEF0CC}0.7037 \\
    \midrule
    \multicolumn{9}{c}{\cellcolor[HTML]{D7D1FF}\textit{Our Generative Reward Models}} \\
    \midrule
    \textbf{RM-NLHF-Qwen-7B} & \textbf{0.7381} & \textbf{0.5757} & \textbf{0.6822} & \textbf{0.6926} & \textbf{0.6583} & \textbf{0.5416} & \textbf{0.5521} & \cellcolor[HTML]{FEF0CC}\textbf{0.6481} \\
    \textbf{RM-NLHF-Qwen-32B} & \textbf{0.8315} & \textbf{0.7867} & \textbf{0.7888} & \textbf{0.7165} & \textbf{0.7492} & \textbf{0.6161} & \textbf{0.6183} & \cellcolor[HTML]{FEF0CC}\textbf{0.7296} \\
    \bottomrule
\end{tabular}}
\end{table}

Table~\ref{tab:main_results} presents the main results across multiple benchmarks, from which we observe:

(1) Compared to specialized trained GRMs with outcome-only reward sharing the same base model, RM-NLHF achieves state-of-the-art performance, demonstrating its effectiveness in producing accurate reward. Specifically, RM-NLHF-Qwen-7B attains an overall score of 0.6481, substantially outperforming RM-R1-Qwen-7B (0.5759) and RRM-Qwen-7B (0.5895). 

(2) Comparing generative reward models with scalar reward models, we find that scalar reward models exhibit superior performance at the 7/8B scale. However, we observe that even when scaling up to 70B, scalar reward models show marginal performance gains. In contrast, generative reward models demonstrate substantial improvements from 7B to 32B. This indicates that generative reward models scale more effectively with model size, while scalar reward models exhibit limited scalability.

\subsection{Ablation Study}

In this section, we provide the main ablation study results in Table~\ref{tab:main_ablation_study}, addressing the following questions.

\noindent\textbf{Ablation on Process Reward.} We train a comparison model by removing process reward while keeping all other configurations (training parameters and data) same. The results show that incorporating process reward yields improvements across all benchmarks, validating the effectiveness and generalizability of process reward. Since we leverage human critiques from HelpSteer3 and generalize to other datasets, this verifies the transferability of HelpSteer3's human critiques. 

\begin{figure}[t]
    \centering
    \begin{subfigure}[t]{0.46\textwidth}
        \centering
        \includegraphics[width=\textwidth]{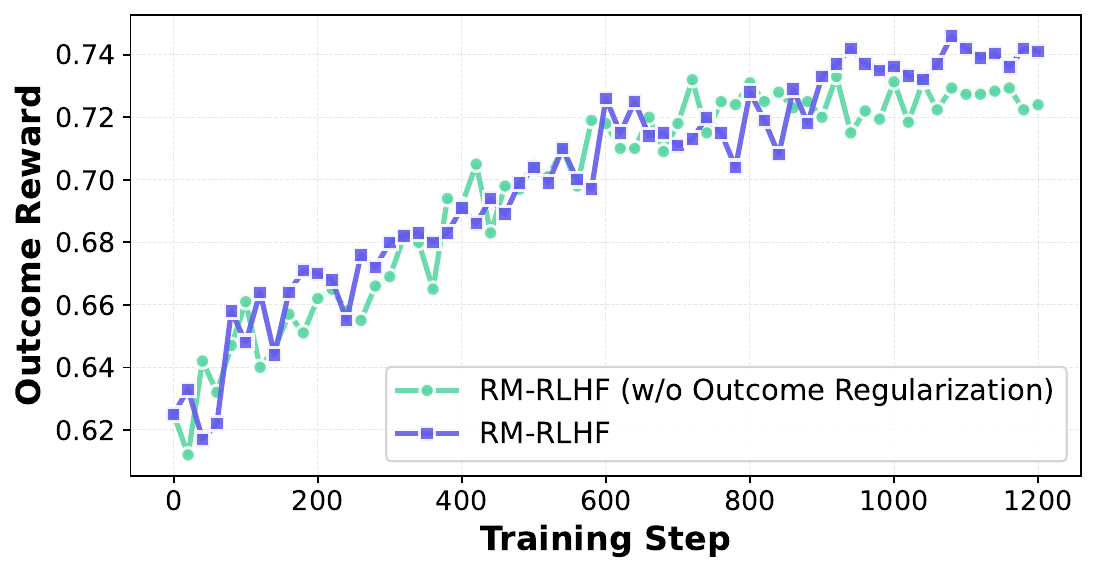}
        \caption{Outcome accuracy on HelpSteer3.}
        \label{fig:val_reward_outcome}
    \end{subfigure}
    \hspace{1.0em}
    \begin{subfigure}[t]{0.47\textwidth}
        \centering
        \includegraphics[width=\textwidth]{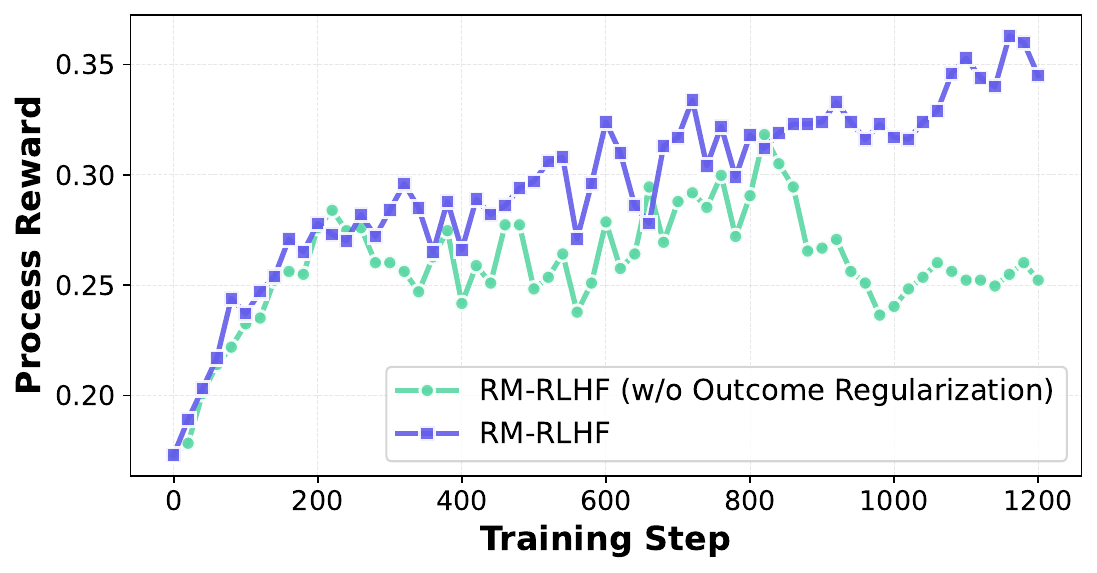}
        \caption{F1-based critique similarity on HelpSteer3.}
        \label{fig:val_critique_score}
    \end{subfigure}
    \caption{Training dynamics of RM-NLHF with and without outcome regularization.}
    \label{fig:val_score}
\end{figure}

\noindent\textbf{Ablation on Outcome Regularization.} In this part, we try to assign process reward even when the outcome is incorrect. However, the results show a substantial decrease in accuracy, which validates the importance of outcome regularization. We visualize the training process in Figure~\ref{fig:val_critique_score}, which demonstrates that both models are initially consistent, but the model without outcome regularization shows stagnant or even declining reward in later stages. This confirms the importance of outcome regularization for maintaining stable long-term RL training. 

\noindent\textbf{Generalizability of Human Critiques in HelpSteer3.} Since our training data combines HelpSteer3 with data from various other sources, we verify whether the human critiques from HelpSteer3 can transfer to other datasets. Specifically, we attempt to also use $D_{o}$ for training MetaRM, where samples with correct and incorrect outcomes are assigned training labels of 0 and $1+\lambda/2$, respectively. This aims to adapt MetaRM to critique distributions beyond HelpSteer3. However, we find that training MetaRM solely on HelpSteer3 yields better performance, indicating that HelpSteer3 provides broad coverage and sufficient transferability for training an effective MetaRM.

\begin{table}[t]
\centering
\small
\caption{Main Ablation Study.}
\resizebox{\textwidth}{!}{
\begin{tabular}{lcccccccc}
\hline
\textbf{Model} & \textbf{HS3} & \textbf{RB-V2} & \textbf{SCAN} & \textbf{HREF} & \textbf{LB} & \textbf{WQA} & \textbf{WPB} & \textbf{AVG} \\
\hline
RM-NLHF & 0.7381 & 0.5757 & 0.6822 & 0.6926 & 0.6583 & 0.5416 & 0.5521 & 0.6481 \\
\hline
\quad w/o Process Reward & 0.7125 & 0.5440 & 0.6709 & 0.6650 & 0.6540 & 0.5314 & 0.5325 & 0.6158 \\
\quad w/o Outcome Regularization & 0.7296 & 0.5529 & 0.6944 & 0.6818 & 0.6512 & 0.5337 & 0.5539 & 0.6282 \\
\quad w/ $D_{o}$ for MetaRM & 0.7291 & 0.5684 & 0.7109 & 0.6857 & 0.6569 & 0.5363 & 0.5456 & 0.6332\\
\hline
\end{tabular}}
\label{tab:main_ablation_study}
\end{table}

\subsection{Performance of Downstream Tasks through Reinforcement Learning}

In this section, we adopt RM-NLHF-Qwen-7B and Outcome-Only-Qwen-7B as the reward models to perform RL training on DeepSeek-Distilled-Qwen-7B. We use prompts from HelpSteer3 as the training set and apply GRPO for 100 training steps. Since both reward models are pairwise in nature, we convert them into pointwise scores via Double-Elimination Tournament~\citep{zhang2026arenarl}.

Results presented in Table~\ref{tab:downstream_rl} demonstrate that our method holds a substantial advantage over the baseline. Across a diverse set of SOTA judge models (gemini-3-pro, gpt-5.1, qwen3-max-thinking and kimi-k2.5~\citep{team2026kimi}), our approach outperforms the baseline by an average margin of 12.92\%. 

\begin{figure}[t]
\centering
    \includegraphics[width=0.8\textwidth]{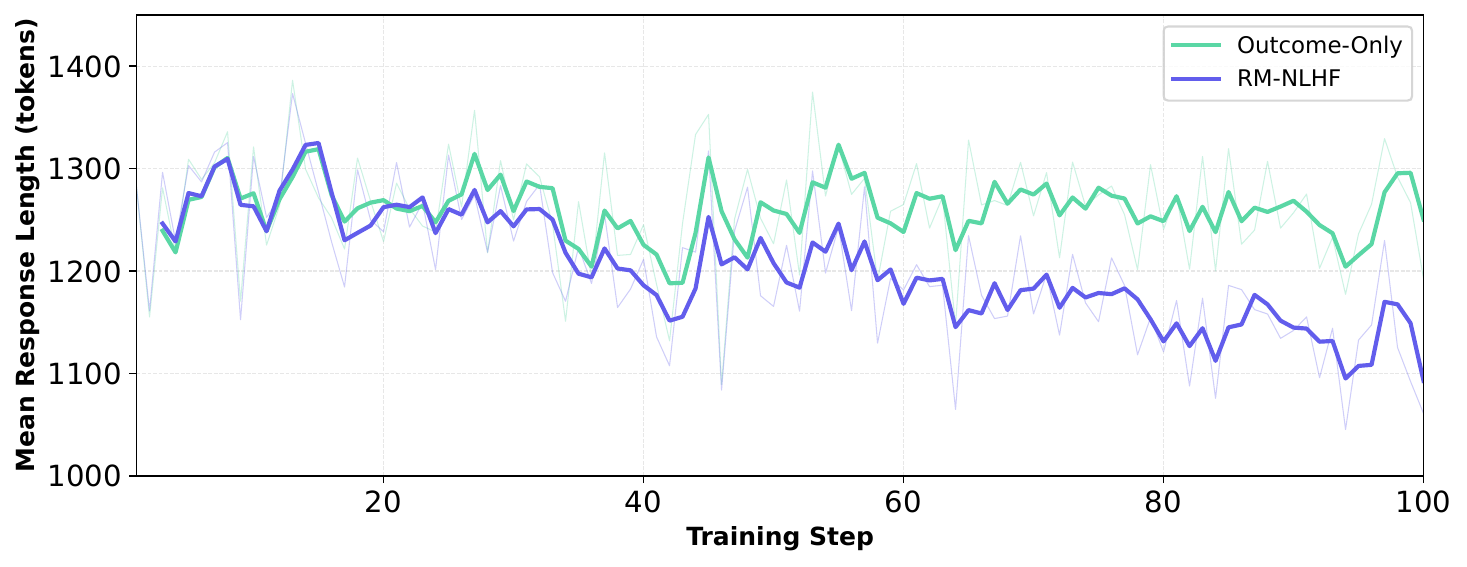}
\caption{Mean response length during downstream RL training.}
\label{fig:downstream_rl_response_length}
\end{figure}

Since GRM is known to exhibit a length bias~\citep{dubois2024length}, we further analyze the output length throughout training, as shown in Figure~\ref{fig:downstream_rl_response_length}. The model trained with RM-NLHF actually produces shorter responses than the baseline, confirming that the gains stem from improved quality rather than verbosity. 

\begin{table}[t]
\centering
\caption{Win rates (\%) of RL-trained models using RM-NLHF vs.\ Outcome-Only as the reward model, evaluated on Arena-Hard-V2.0~\citep{li2025crowdsourced} with different judge models.}
\label{tab:downstream_rl}
\resizebox{0.9\textwidth}{!}{
\begin{tabular}{l cccc c}
\toprule
 & \multicolumn{4}{c}{\textbf{Judge Model}} & \\
\cmidrule(lr){2-5}
 & \textbf{gemini-3-pro} & \textbf{gpt-5.1} & \textbf{qwen3-max-thinking} & \textbf{kimi-k2.5} & \textbf{Avg.} \\
\midrule
RM-NLHF Win      & 57.70 & 54.96 & 49.47 & 47.19 & \cellcolor[HTML]{FEF0CC}\textbf{52.33} \\
Outcome-Only Win  & 34.40 & 37.67 & 43.98 & 41.58 & \cellcolor[HTML]{FEF0CC}39.41 \\
Tie               &  7.90 &  7.37 &  6.55 & 11.23 & \cellcolor[HTML]{FEF0CC}8.26 \\
\bottomrule
\end{tabular}
}
\end{table}

\begin{table}[H]
\centering
\caption{Evaluation results on downstream tasks through test-time scaling.}
\label{tab:bon_critique}
\resizebox{0.7\textwidth}{!}{
\begin{tabular}{l|ccc}
\toprule
    \textbf{Method} & \textbf{MATH500} & \textbf{HumanEval+} & \textbf{Arena-Hard-V2.0} \\
    \midrule
    \multicolumn{4}{l}{\cellcolor[HTML]{EFEFEF}\textbf{Base Model}} \\
    \midrule
    DeepSeek-Distilled-Qwen-7B & 62.92\% & 77.13\% & 3.39\% \\
    \midrule
    \multicolumn{4}{l}{\cellcolor[HTML]{FEF0CC}\textbf{Best-of-N (BoN)}} \\
    \midrule
    Outcome-only (BoN@2) & 63.65\% & 76.30\% & 3.69\% \\
    RM-NLHF (BoN@2) & 64.90\% & 76.95\% & 3.56\% \\
    Outcome-only (BoN@4) & 65.45\% & 75.77\% & 3.93\% \\
    RM-NLHF (BoN@4) & 66.80\% & 81.04\% & 3.85\% \\
    Outcome-only (BoN@8) & 65.99\% & 75.00\% & 4.30\% \\
    RM-NLHF (BoN@8) & \textbf{67.60\%} & \textbf{85.98\%} & \textbf{4.64\%} \\
    \midrule
    \multicolumn{4}{l}{\cellcolor[HTML]{D7D1FF}\textbf{Feedback-Edit}} \\
    \midrule
    Outcome-only & 67.01\% & 82.32\% & 6.55\% \\
    RM-NLHF & \textbf{68.40\%} & \textbf{87.20\%} & \textbf{7.03\%} \\
\bottomrule
\end{tabular}
}
\end{table}

\subsection{Performance of Downstream Tasks through Test-time Scaling}

To verify the effectiveness of GRMs on downstream tasks, beyond following prior work~\cite{rrm} using Best-of-N (BoN), we additionally evaluate the quality of GRM-generated critiques through a Feedback-Edit approach. 
For BoN, we adopt a tournament-based approach where the pairwise GRM selects the best response from N responses sampled from the base model. 
For Feedback-Edit, we use RM-NLHF to select the top 2 responses, then apply GRMs to generate critiques. An edit-model (gemini-2.5-pro) subsequently synthesizes a new response based on these critiques, with the prompt explicitly requiring modifications guided solely by the critiques (see prompt in Figure~\ref{fig:prompt_edit_model}). 

As shown in Table~\ref{tab:bon_critique}, RM-NLHF consistently outperforms the outcome-only baseline across MATH500~\citep{lightman2023let}, HumanEval+~\citep{liu2023your}, and Arena-Hard-V2.0~\citep{li2025crowdsourced}. In the BoN setting, RM-NLHF demonstrates superior ranking capability, with particularly notable improvements on HumanEval+. In the Feedback-Edit setting, RM-NLHF achieves substantial gains over the baseline, validating that process reward enable the generation of higher-quality critiques for response refinement. 

\begin{figure}[H]
    \centering
    \begin{subfigure}[t]{0.46\textwidth}
        \centering
        \includegraphics[width=\textwidth]{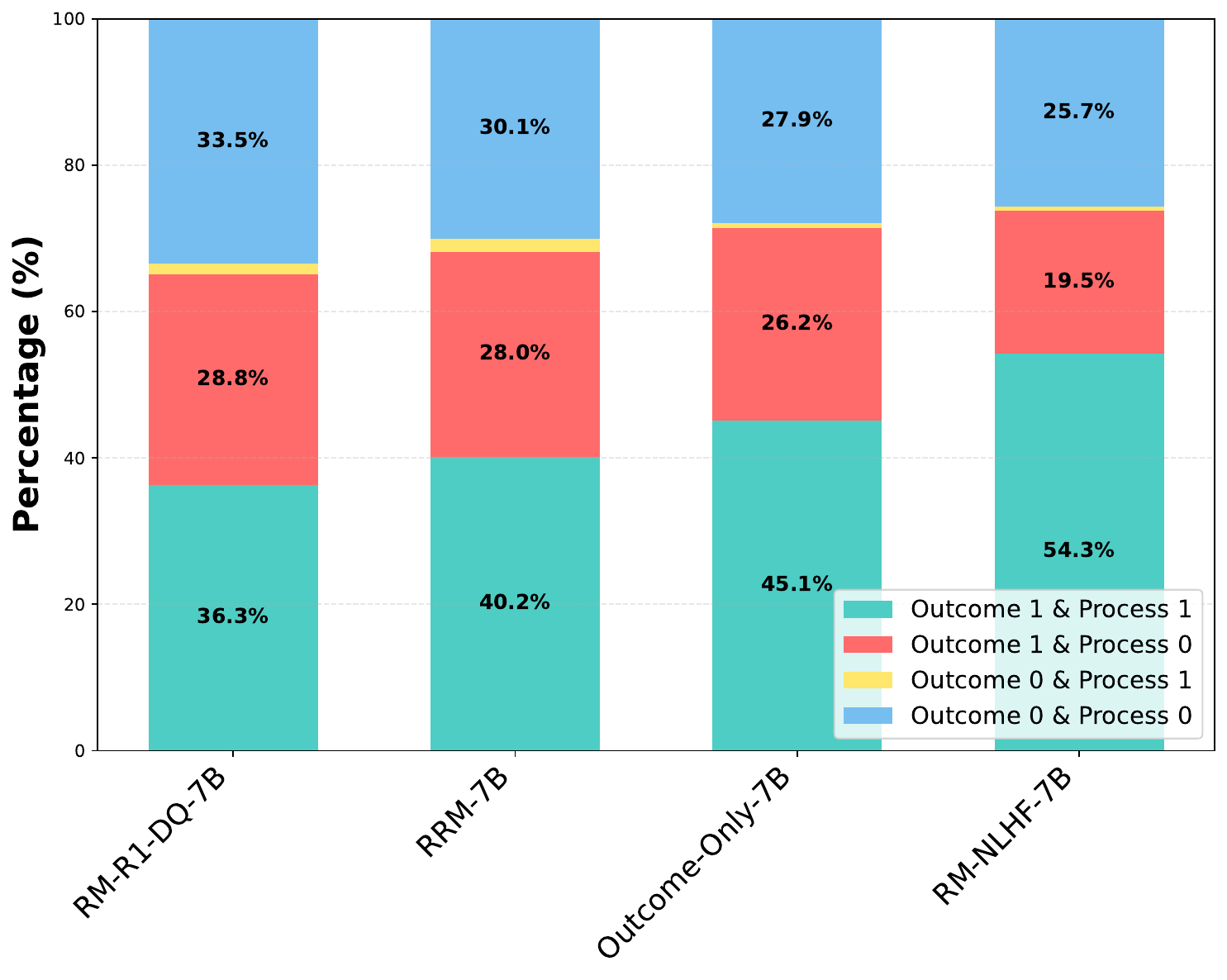}
        \caption{Effectiveness of process reward. $P(\text{process}=0|\text{outcome}=1)$ is substantially lower than baselines.}
        \label{fig:inconsistency_final}
    \end{subfigure}
    \hspace{1.0em}
    \begin{subfigure}[t]{0.47\textwidth}
        \centering
        \includegraphics[width=\textwidth]{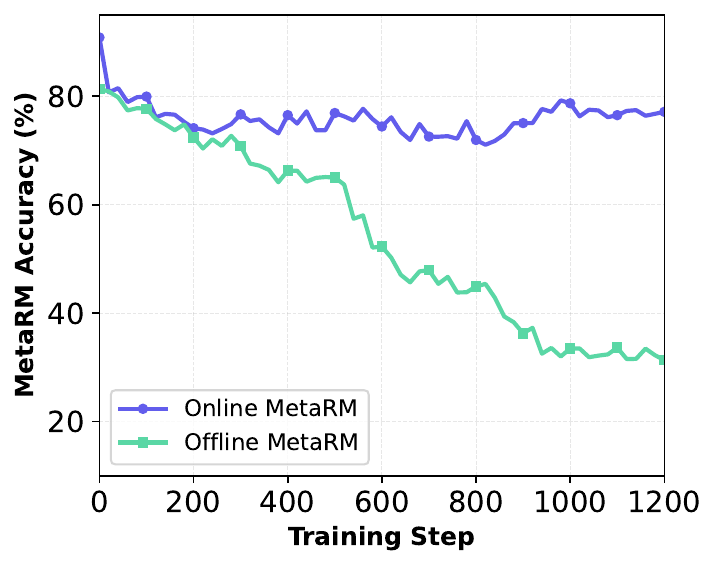}
        \caption{Effectiveness of online updating. Online MetaRM maintains higher accuracy than offline MetaRM.}
        \label{fig:acc_online_offline_metarm}
    \end{subfigure}
    \caption{Effectiveness of human critiques and online MetaRM as process reward.}
    \label{fig:online_offline_metarm_acc_comparison}
\end{figure}

\subsection{Analysis of Computational Cost}

An important question is how much additional time overhead our method introduces. We provide a detailed breakdown of the time consumption for one RL training step in Table~\ref{tab:time_breakdown}. 
Our scoring scheme can be executed either synchronously or asynchronously with the rollout process. In the synchronous setting (Ours(Sync)), our method increases the training time from 156s to 289s for the 7B model and from 655s to 874s for the 32B model per training step. However, with asynchronous execution (Ours(Async)), the overhead is significantly reduced: the total time becomes 196s for 7B (26\% increase) and 766s for 32B (17\% increase) compared to the baseline. 
Considering the significant improvement in final performance, the additional computation overhead (26\% for 7B and 17\% for 32B) introduced by our method is acceptable.

\begin{table*}[t]
\centering
\caption{\textbf{Training Time Breakdown for One Step (seconds). Processes with the same color can be executed asynchronously.} We enable asynchronous execution for computing reward and computing log-probabilities by default. "Ours(Sync)" and "Ours(Async)" denote sequential and concurrent execution of Rollout and Rewarding w/ HC (Human Critique) \& Outcome, respectively.}
\label{tab:time_breakdown}
\resizebox{\textwidth}{!}{
    \begin{tabular}{lcccccccc}
    \hline
    \textbf{Method} & \textbf{Rollout} & \textbf{\makecell{Rewarding \\ w/ HC \& Outcome}} & \textbf{\makecell{MetaRM \\ (Update \& Rewarding)}} & \textbf{\makecell{Compute \\ LogProb \& KL}} & \textbf{\makecell{Update \\ Policy}} & \textbf{\makecell{Save \\ Logs}} & \textbf{Total} \\
    \hline
    \multicolumn{8}{l}{\textit{7B}} \\
    \hline
    Outcome-only & \cellcolor{blue!20}95 & \cellcolor{green!20}2 & - & \cellcolor{green!20}21 & \cellcolor{red!20}36 & \cellcolor{orange!20}4 & \textbf{156} \\
    Ours (Sync) & \cellcolor{blue!20}95 & \cellcolor{green!20}113 & \cellcolor{cyan!20}41 & \cellcolor{cyan!20}21 & \cellcolor{red!20}36 & \cellcolor{orange!20}4 & \textbf{289} \\
    Ours (Async) & \cellcolor{blue!20}95 & \cellcolor{blue!20}115 & \cellcolor{cyan!20}41 & \cellcolor{cyan!20}21 & \cellcolor{red!20}36 & \cellcolor{orange!20}4 & \textbf{196} \\
    \hline
    \multicolumn{8}{l}{\textit{32B}} \\
    \hline
    Outcome-only & \cellcolor{blue!20}376 & \cellcolor{green!20}2 & - & \cellcolor{green!20}91 & \cellcolor{red!20}184 & \cellcolor{orange!20}4 & \textbf{655} \\
    Ours (Sync) & \cellcolor{blue!20}376 & \cellcolor{green!20}113 & \cellcolor{cyan!20}197 & \cellcolor{cyan!20}91 & \cellcolor{red!20}184 & \cellcolor{orange!20}4 & \textbf{874} \\
    Ours (Async) & \cellcolor{blue!20}376 & \cellcolor{blue!20}381 & \cellcolor{cyan!20}197 & \cellcolor{cyan!20}91 & \cellcolor{red!20}184 & \cellcolor{orange!20}4 & \textbf{766} \\
    \hline
    \end{tabular}}
\end{table*}

\subsection{How Do Human Critiques Take Effect?}

To demonstrate how human critiques improve critique quality, we present the proportion of cases where our model produces correct outcomes but flawed processes in Figure~\ref{fig:inconsistency_final}. As shown, compared to similar-sized models, our method reduces $P(\text{process}=0|\text{outcome}=1)$ significantly, demonstrating the superiority of our approach in generating higher-quality critiques.

\subsection{How Does Online MetaRM Updating Take Effect?}

To demonstrate the importance of online MetaRM training, we evaluate MetaRM's accuracy throughout the training process using the validation set from HelpSteer3. Accuracy is defined as the agreement between MetaRM-assigned scores and those from the F1-based similarity between critiques (gemini-2.5-pro). We report the results in Figure~\ref{fig:acc_online_offline_metarm}, which clearly demonstrates that online updates maintain high accuracy by continuously adapting to the evolving policy model's output distribution. In contrast, while the offline MetaRM starts with comparable accuracy, its performance gradually deteriorates throughout RL training, highlighting the critical advantage of online training for maintaining reward model quality during iterative long-term RL optimization.

\subsection{Critique Evolution: From Binary to Natural Language Feedback}
\label{sec:ablation_on_critique_evolution}

To examine how RM-NLHF transforms critique generation capability compared to baseline model (trained with same configuration except using outcome reward), we conduct linguistic analysis on the HelpSteer3 validation set. 

\paragraph{Vocabulary Diversity.}
We first examine how many unique adjectives and verbs are included in their generated critiques. The results show that RM-NLHF produces critiques with \textbf{4,436} unique words, compared to only \textbf{1,633} for the baseline. This substantial improvement indicates RM-NLHF generates more specific, targeted and diverse feedback, avoiding superficial, repetitive, templated patterns. 

\begin{figure}[H]
\centering
    \includegraphics[width=0.7\linewidth]{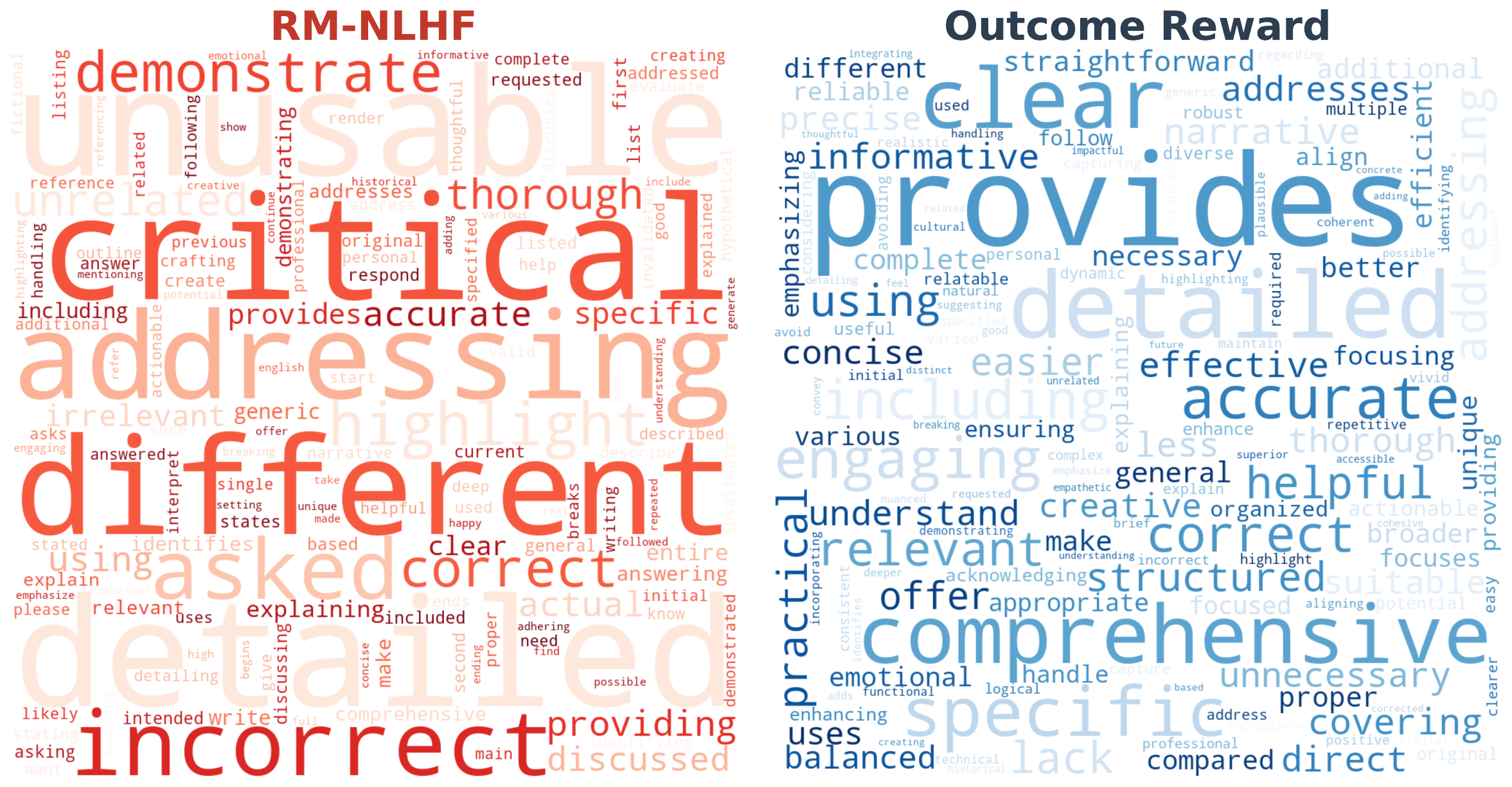}
\caption{Word cloud comparison of RM-NLHF vs. baseline critiques.}
\label{fig:critique_wordcloud}
\end{figure}

\paragraph{Word Frequency Analysis.}
We present the word frequency differences in Table~\ref{tab:characteristic_terms} and visualize them through word clouds in Figure~\ref{fig:critique_wordcloud}. The results show that RM-NLHF employs critical and diagnostic terms like ``unusable'' (+0.033), ``incorrect'' (+0.022), and ``unrelated'' (+0.010), reflecting targeted, actionable feedback. In contrast, the baseline over-relies on generic positive descriptors like ``comprehensive'' (-0.029), ``clear'' (-0.029), and ``helpful'' (-0.007), suggesting superficial praise over substantive evaluation. 

Overall, this linguistic analysis reveals a fundamental limitation of pure outcome-based reward models: they tend to generate superficial, templated and nitpicky feedback that fails to capture the true core issues. 
However, by incorporating explicit natural language feedback signals, we force the model to identify the key factors that truly distinguish response quality, thereby encouraging the model to produce critiques that are more precise and diverse. 

\begin{table}[H]
\centering
\caption{Distinguishing terms between RM-NLHF and baseline critiques.}
\label{tab:characteristic_terms}
\small
\resizebox{0.67\textwidth}{!}{
\begin{tabular}{ll|ll}
\toprule
\multicolumn{2}{c|}{\textbf{RM-NLHF}} & \multicolumn{2}{c}{\textbf{Baseline (Outcome Reward)}} \\
\textbf{Term} & \textbf{Freq. Diff.} & \textbf{Term} & \textbf{Freq. Diff.} \\
\midrule
unusable & +0.033 & provides & -0.089 \\
critical & +0.030 & comprehensive & -0.029 \\
different & +0.027 & clear & -0.029 \\
incorrect & +0.022 & specific & -0.015 \\
asked & +0.020 & engaging & -0.014 \\
addressing & +0.018 & detailed & -0.010 \\
highlights & +0.013 & including & -0.009 \\
demonstrates & +0.012 & lacks & -0.009 \\
unrelated & +0.010 & helpful & -0.007 \\
actual & +0.009 & practical & -0.007 \\
\bottomrule
\end{tabular}
}
\end{table}

%% file: content/related_work.tex
\section{Related Work}

Generative reward models (GRMs) have emerged as a powerful alternative to traditional scalar reward models~\citep{ouyang2022training,sun2025rethinking,zhang2024generative}, offering greater interpretability and accuracy by generating natural language critiques and reasoning chains.
The LLM-as-a-Judge paradigm~\citep{zheng2023judging,dubois2024length,saha2023branch} pioneered this approach, producing explicit rationales~\citep{kim2024prometheus,ankner2024critique,yu2024self,saha2025learning} that enhance both transparency and evaluation accuracy.
Training methods have evolved from prompt engineering to supervised fine-tuning (SFT) and direct preference optimization (DPO)~\citep{ye2024beyond,wang2024direct,mahan2024generative,ye2024improving,wu2024meta,zhao2025genprm,anugraha2025r3}, and more recently to reinforcement learning with verifiable reward (RLVR)~\citep{rmr1,rrm,yang2025deepcritic,xu2025j4r,yu2025rewardanything}, inspired by advances in reasoning models~\citep{r1,team2025kimi,hu2025open,xie2025logic}.

%% file: content/conclusion.tex
\section{Conclusion}

In this work, we identified and addressed the outcome-process inconsistency problem in generative reward models (GRMs). We incorporated human critiques as process reward to improve critique quality and preference accuracy. To address scalability challenges, we proposed MetaRM, a meta reward model that learns from limited human critiques and generalizes to unlabeled data. We further introduced an online MetaRM framework that adapts to distribution shifts during training. Experiments show our method achieves performance higher that outcome-only supervision.

%% file: content/appendix.tex
\appendix
\clearpage

\section{Discussion about Reward Hacking}
\label{appendix:discussion_reward_hacking}


During our experiments, we observed multiple forms of reward hacking. We document these engineering insights here as a discussion of the reward hacking phenomenon. 

\noindent\textbf{Reward Hacking for Recall-based Similarity.} 
Initially, we attempted to use recall to measure the similarity between GRM-generated critiques and human critiques. However, we found that the GRM learned to generate an excessive number of critiques to maximize recall. Although it increased the training reward, it led to degraded performance on benchmarks. 

\noindent\textbf{Reward Hacking for Process Reward.} 
We experimented with F1-based similarity and MetaRM and training for an extended number of steps. However, in later stages of training, the model exhibited abnormal behavior: response lengths became excessively long, process reward increased unreasonably, and outcome reward declined. Specifically, the GRM learned to exploit the system by repeatedly generating identical critiques. Therefore, in the prompt for computing similarity between critiques, we instructed it to first check whether the model repeats the same arguments multiple times. 

\section{Discussion about Future Work}
\label{appendix:discussion_future_work}

Although our method was validated on GRMs, it fundamentally addresses the challenge of obtaining process-level reward. We believe our approach can be extended to two promising domains.

\noindent\textbf{Multiple-Choice and True/False Questions.}
First, tasks such as multiple-choice or true/false questions naturally suffer from noisy outcome reward due to their restricted solution space. Our method could provide more informative process-level supervision in these settings.

\noindent\textbf{Open-Ended Tasks with Verifiable Correctness.}
Second, our approach may benefit open-ended tasks that lack explicit outcome reward but have verifiable correctness criteria, such as mathematical proofs. In these cases, correctness can be determined by comparing against ground truth solutions, providing process-level supervision signals.

\section{Discussion about Limitations}
\label{appendix:limitation}

We discuss the limitations of our approach here. 

\noindent\textbf{Limitations on Fully Open-Ended Tasks.}
We note that our method may not directly transfer to fully open-ended tasks without any ground truth, as these lack the human critiques or verifiable signals that our framework relies upon. Extending to such scenarios remains an important direction for future research. 

\noindent\textbf{Dependency on High-Quality Human Critiques.} Our method requires high-quality human critiques. Although MetaRM enables semi-supervised learning and knowledge transfer, the approach may face challenges when dealing with diverse preferences across different populations, such as regional or cultural variations. In such cases, critiques from HelpSteer3 may not transfer effectively to these datasets. In this work, we focus on general-purpose scenarios where HelpSteer3 critiques are broadly applicable. 

\noindent\textbf{API Cost from External Models.} We need to invoke external models to compute similarity between GRM-generated critiques and human critiques, which introduces additional costs. This is primarily due to the limited capabilities of our 7B and 32B models. For larger models with stronger capabilities, similarity computation could be performed internally, eliminating this overhead. This self-verification approach (such as \citet{shao2025deepseekmath}) represents a promising direction for future work. 

\newpage
\section{MetaRM}
\label{appendix:metarm}

\subsection{Online Training Algorithm}
\label{appendix:online_metarm}

The online MetaRM framework alternates between updating the policy GRM $\pi_\theta$ and the MetaRM $M_\phi$, ensuring that the reward model remains aligned with the evolving generation distribution. Algorithm~\ref{alg:online_metarm} presents the complete training procedure.

\begin{algorithm}[t]
\caption{Online MetaRM Training for GRM}
\label{alg:online_metarm}
\begin{algorithmic}[1]
\Require Dataset with human critiques $\mathcal{D}_H = \{(q, y_A, y_B, l, h)\}$, dataset without critiques $\mathcal{D}_O = \{(q, y_A, y_B, l)\}$, initial GRM $\pi_{\theta_0}$, initial MetaRM $M_{\phi_0}$, process reward weight $\lambda$, rollout budget $N_{\text{rollout}}$, pre-training epochs $E_{\text{pre}}$

\State \textcolor{blue}{// Cold start: Initialize MetaRM}
\For{epoch $e = 1, 2, \ldots, E_{\text{pre}}$}
    \For{each sample $(q, y_A, y_B, h) \in \mathcal{D}_H$}
        \State Sample $N_{\text{rollout}}$ responses $\{\hat{y}_i\}_{i=1}^{N_{\text{rollout}}}$ from $\pi_{\theta_0}(q)$
        \State Evaluate critique quality: $R_{\text{process}} = \mathbb{I}[S(h, \hat{c}) > 0.5]$
        \State Compute target reward $R_{\text{target}}$ using Equation~\ref{eq:composite_reward}
    \EndFor
    \State Update MetaRM ${M_{\phi}}_{\text{cold}}$ using MSE loss (Equation~\ref{eq:metarm_loss})
\EndFor

\State
\State \textcolor{blue}{// GRPO with online MetaRM}
\For{step $t = 1, 2, \ldots, T$}
    \State \textcolor{blue}{// Step 1: Generate rollouts from current policy}
    \For{each query $q$ in minibatch}
        \State Sample $N_{\text{rollout}}$ responses $\{\hat{y}_i\}_{i=1}^{N_{\text{rollout}}}$ from $\pi_{\theta}(q)$
    \EndFor
    
    \State \textcolor{blue}{// Step 2: Compute reward for $\mathcal{D}_H$ using human critiques}
    \For{each sample $(q, y_A, y_B, h) \in \mathcal{D}_H$}
        \State Evaluate critique quality: $R_{\text{process}} = \mathbb{I}[S(h, \hat{c}) > 0.5]$
        \State Compute reward $R$ using Equation~\ref{eq:composite_reward}
    \EndFor
    
    \State \textcolor{blue}{// Step 3: Update MetaRM using $\mathcal{D}_H$ and reward $R$}
    \State Update MetaRM using Equation~\ref{eq:metarm_loss}
    
    \State \textcolor{blue}{// Step 4: Compute reward for $\mathcal{D}_O$ using updated MetaRM}
    \For{each sample $(q, y_A, y_B) \in \mathcal{D}_O$}
        \State Predict meta-reward $\hat{R}_{\text{meta}}$ using Equation~\ref{eq:infer_metarm}
        \State Assign reward $R'$ for this sample using Equation~\ref{eq:metarm_reward}
    \EndFor
    
    \State \textcolor{blue}{// Step 5: Update policy model using reward from both datasets}
    \State Compute advantages $\hat{A}_i$ for all rollouts using reward $\{R, R'\}$
    \State Update policy model (GRM) with GRPO loss using Equation~\ref{eq:grpo_loss}
\EndFor
\end{algorithmic}
\end{algorithm}

\paragraph{Key Design Principles.}
The online training procedure incorporates several important design choices:

\begin{itemize}[leftmargin=*]
    \item \textbf{Dual Data Streams}: We maintain two data streams: $\mathcal{D}_H$ with human critiques for MetaRM supervision, and $\mathcal{D}_O$ without critiques for scalable policy learning.
    
    \item \textbf{MetaRM-First Update}: In each iteration, we first update the MetaRM using fresh samples from $\mathcal{D}_H$ (Step 3), then use the updated MetaRM to score samples in $\mathcal{D}_O$ (Step 4). This ensures the MetaRM is fully on-policy before providing reward. 
    
    \item \textbf{Continuous Adaptation}: Unlike offline training where the MetaRM is frozen after initialization, online training continuously adapts the MetaRM to the evolving critique distribution as the policy improves.
\end{itemize}

\section{Experiment Setups for Exploring Process Reward Design}
\label{appendix:exploring_process_reward_design_for_grm_training}

For dataset construction, we randomly sample 49 data instances from HelpSteer3, each containing a query $q$, two responses $y_a$ and $y_b$, GRM-generated critiques $\hat{c}$, and human critiques $h$. We manually annotate each critique $\hat{c}$ with a binary label $z \in \{0, 1\}$, where $z=1$ indicates the critique is correct (i.e., accurately identifies issues in the response), and $z=0$ otherwise. This results in a small but carefully curated evaluation set for studying process supervision methods. 

We investigate three approaches to automatically predict the correctness label:

\paragraph{(1) LLM-as-a-Meta-Judge.} We prompt an external LLM to directly evaluate whether the critique $\hat{c}$ correctly identifies issues in the response, given the query $q$ and response $y$. The prompt is shown in Figure~\ref{fig:prompt_llm_as_a_meta_judge}. 

\paragraph{(2) Similarity w/ All HC.} We use an external LLM to extract all critique points from both $h$ and $\hat{c}$, then compute their similarity using three metrics: F1, Recall, and Precision. High similarity suggests the GRM critique aligns with human judgment. The extraction prompt is shown in Figure~\ref{fig:prompt_similarity_core_hc}.

\paragraph{(3) Similarity w/ Core HC.} Similar to (2), but we prompt the LLM to identify and extract only the \emph{core} critical points from $h$ and $\hat{c}$, filtering out minor issues. We then compute similarity between these core critiques. The extraction prompt is shown in Figure~\ref{fig:prompt_similarity_core_hc}.

\section{Experiment Setups for Preliminary Results}
\label{appendix:exp_preliminary_details}

\noindent\textbf{Experiment Setups for Section~\ref{sec:human_critique_process_reward_integration}.}
For training hyperparameters, we set the training steps to 150 (1 epoch), while all other parameters follow Table~\ref{tab:training_hyperparameters}. We employ gpt-5-mini as the scoring model to evaluate the similarity (F1 score) between human critiques and scores. 
For training data, we use the complete HelpSteer3 dataset. 

\noindent\textbf{Experiment Setups for Section~\ref{sec:method_metarm}.}
For training hyperparameters, we set the training steps to 150 (1 epoch), while all other parameters follow Table~\ref{tab:training_hyperparameters}. We employ gpt-5-mini as the scoring model to evaluate the similarity (F1 score) between human critiques and GRM-generated critiques. 
For training data, we randomly split the HelpSteer3 dataset into two equal parts with a 1:1 ratio, serving as $\mathcal{D}_H$ and $\mathcal{D}_O$, respectively.

\section{Experiment Setups for Main Results}
\label{appendix:exp_details}

\subsection{Benchmarks}

Following previous work~\citep{shao2025spurious}, data contamination has a significant impact on training effectiveness and evaluation credibility, especially for the Qwen2.5 series of models, where sometimes even spurious reward can take effect. In this paper, we mitigate this issue by selecting benchmarks that are released at most 1 month before the base model. 

\begin{table}[h]
\centering
\caption{Base model and benchmark release timeline, domains, label sources, and usage. Benchmarks released within 1 month before the base model are excluded to mitigate data contamination concerns.}
\label{tab:benchmarks}
\scriptsize
\begin{tabular}{m{3.5cm}m{1.2cm}m{4.5cm}m{3cm}m{0.8cm}}
\hline
\textbf{Benchmark} & \textbf{Release} & \textbf{Domain} & \textbf{Label Source} & \textbf{Used} \\
\hline
DeepSeek-R1-Distill-Qwen-7B~\citep{r1} & 2025.01 & -- & -- & -- \\ \hline
DeepSeek-R1-Distill-Llama-8B~\citep{r1} & 2025.01 & -- & -- & -- \\ \hline
DeepSeek-R1-Distill-Qwen-32B~\citep{r1} & 2025.01 & -- & -- & -- \\ \hline \hline
MT-Bench~\citep{zheng2023judging} & 2023.06 & Writing, Roleplay, Extraction, Reasoning, Math, Coding, STEM, Humanities & Majority human voting & \ding{55} \\ \hline
RewardBench~\citep{lambert2025rewardbench} & 2024.03 & Chat, Safety, Math, Code & Verifiable correctness, dataset integration & \ding{55} \\ \hline
RM-Bench~\citep{liu2024rm} & 2024.10 & Chat, Safety, Code, Math & Verifiable correctness, deliberate errors & \ding{55} \\ \hline
PPE~\citep{frick2025evaluate} & 2024.10 & Knowledge, Math, STEM, Coding, Instruction Following, Chat & Verifiable correctness, single human & \ding{55} \\ \hline
JudgeBench~\citep{tan2024judgebench} & 2024.10 & Knowledge, Reasoning, Math, Coding & Verifiable correctness & \ding{55} \\ \hline
RMB~\citep{zhou2024rmb} & 2024.10 & Helpfulness, Harmlessness & gpt-4o & \ding{55} \\ \hline
HREF~\citep{lyu2024href} & 2024.12 & General instruction following & Majority human voting & \checkmark \\ \hline
WQ-LMArena~\citep{chakrabarty2025ai} & 2025.04 & General writing & Single human & \checkmark \\ \hline
SCAN-HPD~\citep{wang2025scan} & 2025.05 & Writing, Roleplay, Knowledge, Coding, Math, Reasoning & Majority human voting & \checkmark \\ \hline
HelpSteer3~\citep{wang2025helpsteer3} & 2025.05 & General chat, STEM, Code, Multilingual & Majority human voting & \checkmark \\ \hline
RewardBench V2~\citep{malik2025rewardbench} & 2025.06 & Factuality, Instruction Following, Math, Safety, Focus & LLM-as-a-judge, verifiable correctness, human & \checkmark \\ \hline
LitBench~\citep{fein2025litbench} & 2025.07 & Social media replies & Human upvote count & \checkmark \\ \hline
WPB~\citep{ying2025beyond} & 2025.10 & Documents, Communication, Fiction, Poetry, Scriptwriting, Role-playing & Human-AI collaborative & \checkmark \\ \hline
\end{tabular}
\end{table}

\subsection{Training Datasets}

In our experiments, we exclusively use pairwise preference data. The specific distribution of our training data is shown in Table~\ref{tab:dataset_stats}. 

\begin{table}[h]
\centering
\caption{\textbf{Training Data Overview.} The asterisk (*) denotes data sourced from Skywork-RewardPreference-80K-v0.2~\citep{liu2024skywork}.}
\label{tab:dataset_stats}
\small
\begin{tabular}{lrcc}
\hline
\textbf{Source} & \textbf{Size} & \textbf{Domain} \\
\hline
\multicolumn{4}{c}{\textit{Process Reward: Human Critique}} \\ \hline
    HelpSteer3~\citep{wang2025helpsteer3} & 36,255 & General Domain & \\
\hline
\multicolumn{4}{c}{\textit{Process Reward: MetaRM}} \\ \hline
    Helpsteer2~\citep{wang2024helpsteer}* & 7,221 & General Domain & \\
    Magpile\_Pro\_Llama3.1~\citep{xu2025magpie}* & 28,349 & General Domain & \\
    Magpile\_Pro~\citep{xu2025magpie}* & 2,030 & General Domain & \\
    Offset\_Bias~\citep{park2024offsetbias}* & 8,504 & General Domain & \\
    \href{https://huggingface.co/datasets/Vezora/Code-Preference-Pairs}{Code-Preference-Pairs} & 8,000 & Code & \\
    Math-DPO-10K~\citep{lai2024step} & 10,000 & MATH & \\
    \href{https://huggingface.co/datasets/hamishivi/math_rlvr_mixture_dpo/viewer/}{hamishivi/math\_rlvr\_mixture\_dpo} & 17,713 & MATH & \\
    WebInstruct~\citep{sun2025s2j} & 5,910 & Knowledge & \\
    WildGuard~\citep{han2024wildguard}* & 6,709 & Safety & \\
    \href{https://huggingface.co/datasets/javirandor/hh-rlhf-safety-v3-dpo}{javirandor/hh-rlhf-safety-v3-dpo} & 9,371 & Safety & \\
    Tulu-3-Pref-Personas-Instruction-Following~\citep{lambert2024tulu} & 12,000 & Instruction Following & \\
    LitBench-Train~\citep{fein2025litbench} & 12,000 & Social Media & \\
\hline
\textbf{Total} & \textbf{164,062} & & \\
\hline
\end{tabular}
\end{table}

\subsection{Implementation Details}

For our main experiments, we select Deepseek-R1-Distill-Qwen-7B and Deepseek-R1-Distill-Qwen-32B as the base models. To evaluate the F1-based similarity between generated critiques and human critiques, we employ the lightweight gpt-5-mini as the scoring model. For MetaRM, we utilize the same model as GRM but replace the final layer with a regression head that outputs a single scalar value. 

For the cold start of MetaRM, we first sample 8 responses per prompt from the base model using HelpSteer3. We then train it for 3 epochs with a learning rate of 1e-5, using a cosine scheduler, weight decay of 0.01, and a warmup ratio of 0.01. 

We train our GRMs using the GRPO algorithm, implemented on the VeRL~\citep{sheng2024hybridflow} framework. The training is conducted on a setup with 32 NVIDIA A100 GPUs. Key hyperparameters are detailed in Table~\ref{tab:training_hyperparameters}. 

\begin{table}[t]
\centering
\caption{Training hyperparameters.}
\label{tab:training_hyperparameters}
\resizebox{0.4\columnwidth}{!}{%
\begin{tabular}{lc}
\toprule
    \textbf{Hyperparameter} & \textbf{Value} \\
    \midrule
    \multicolumn{2}{l}{\textit{Data Configuration}} \\
    \midrule
    Max prompt length & 6144 \\
    Max response length & 8192 \\
    \midrule
    \multicolumn{2}{l}{\textit{GRPO Algorithm Configuration}} \\
    \midrule
    Rollouts per prompt & 8 \\
    Sampling temperature & 0.7 \\
    Sampling top-p & 1.0 \\
    KL loss & 0.001 \\
    \midrule
    \multicolumn{2}{l}{\textit{GRM}} \\
    \midrule
    Optimizer & AdamW \\
    Learning rate & 1e-6 \\
    Batch size & 256 \\
    Mini-batch size & 256 \\
    Total training steps & 1200 \\
    \midrule
    \multicolumn{2}{l}{\textit{Online MetaRM during RL training}} \\
    \midrule
    Loss & MSE loss \\
    Learning rate & 5e-6 \\
\bottomrule
\end{tabular}}
\end{table}

\subsection{Baselines}

We select open-source specialized GRMs: RM-R1 and RRM, using their versions based on Deepseek-R1-Distill-Qwen-7B and Deepseek-R1-Distill-Qwen-32B respectively for fair comparison. 
For scalar reward model, we select URM-LLaMa-3.1-8B~\citep{lou2024uncertainty}, Skywork-Reward-Llama-3.1-8B-v0.2~\citep{liu2024skywork}, ArmoRM-Llama3-8B-v0.1~\citep{ArmoRM} and INF-ORM-Llama3.1-70B~\citep{INF-ORM-Llama3.1-70B}. 
Additionally, we employ several base models as judge models through prompt engineering, including: gemini-2.5-pro, claude-3-7-sonnet-20250219, o3-2025-04-16, gpt-5-2025-08-07, gpt-4o-latest, deepseek-r1-0528-inner, deepseek-v3-inner, qwen-plus-latest, DeepSeek-R1-Distill-Qwen-7B, DeepSeek-R1-Distill-Llama-8B, and DeepSeek-R1-Distill-Qwen-32B.

\section{More Results}
\label{appendix:more_results}

\subsection{Ablation Study on MetaRM}
\label{appendix:ablation_on_metarm}

We conduct ablation studies on MetaRM to analyze the impact of different reward designs for the outcome dataset $\mathcal{D}_O$. Note that for the human critique dataset $\mathcal{D}_H$, all variants compute similarity with human critiques. We compare the following reward strategies for $\mathcal{D}_O$:
(1) \textbf{No-MetaRM}: using only outcome-based reward; 
(2) \textbf{Offline-MetaRM}: reward from offline-trained MetaRM with continuous values; 
(3) \textbf{Online-MetaRM}: our default online MetaRM with continuous reward; 
(4) \textbf{Only-MetaRM}: using only MetaRM reward without any outcome signal;
(5) \textbf{Binary}: reward from online MetaRM with outputs discretized to \{0, 1\}; 
(6) \textbf{Aggressive}: online MetaRM trained with higher learning rate (1e-5) for 2 epochs per round; 
(7) \textbf{Classifier}: online MetaRM using classification head instead of regression. 

The ablation results reveal two critical findings. First, online training is essential for MetaRM's effectiveness. Online-MetaRM consistently outperforms Offline-MetaRM across both base models (0.5552 vs 0.5493 for Qwen-7B; 0.6356 vs 0.6273 for Llama-8B), as online adaptation enables MetaRM to track the evolving policy distribution during RL training. The Aggressive variant, which updates MetaRM too rapidly, degrades performance (0.5407 for Qwen-7B), indicating that overly fast adaptation destabilizes training. Second, using MetaRM reward alone (Only-MetaRM) exhibits model-dependent instability. While Only-MetaRM achieves the best results for Qwen-7B (0.5579), it significantly underperforms for Llama-8B (0.6143 vs 0.6356), with a dramatic drop on RewardBench2 (0.3562 vs 0.5196). This inconsistency suggests that relying solely on MetaRM without grounding in outcome-based reward leads to unpredictable behavior across architectures. The default Online-MetaRM, which combines both signals, provides more robust performance. Additionally, Binary and Classifier variants both underperform, confirming that continuous regression-based reward are preferable for capturing nuanced quality distinctions. 

\begin{table}[h]
\centering
\small
\caption{Ablation study on MetaRM designs.}
\label{tab:metarm_ablation}
\resizebox{\textwidth}{!}{
\begin{tabular}{lcccccccc}
\toprule
\textbf{Method} & \textbf{HelpSteer3} & \textbf{RewardBench2} & \textbf{SCAN-HPD} & \textbf{HREF} & \textbf{LitBench} & \textbf{WQ-Arena} & \textbf{WPB} & \textbf{Overall} \\
\midrule
\multicolumn{9}{l}{\textit{DeepSeek-R1-Distill-Qwen-7B}} \\
\midrule
No-MetaRM & 0.6242 & 0.3375 & 0.6245 & 0.5603 & 0.5392 & 0.5329 & 0.5240 & 0.5347 \\
Offline-MetaRM & 0.6444 & 0.3432 & 0.6440 & 0.5628 & 0.5680 & 0.5310 & 0.5516 & 0.5493 \\
Online-MetaRM & 0.6529 & 0.3364 & 0.6712 & 0.5719 & 0.5738 & 0.5335 & 0.5464 & 0.5552 \\
Only-MetaRM & 0.6723 & 0.3358 & 0.6840 & 0.5554 & 0.5686 & 0.5309 & 0.5582 & \textbf{0.5579} \\
Binary & 0.6615 & 0.3307 & 0.6342 & 0.5661 & 0.5662 & 0.5359 & 0.5335 & 0.5469 \\
Aggressive & 0.6294 & 0.3358 & 0.6407 & 0.5504 & 0.5534 & 0.5289 & 0.5464 & 0.5407 \\
Classifier & 0.6382 & 0.3301 & 0.6384 & 0.5421 & 0.5647 & 0.5348 & 0.5315 & 0.5400 \\
\midrule
\multicolumn{9}{l}{\textit{DeepSeek-R1-Distill-Llama-8B}} \\
\midrule
No-MetaRM & 0.7096 & 0.5054 & 0.7396 & 0.5901 & 0.5919 & 0.5651 & 0.6156 & 0.6168 \\
Offline-MetaRM & 0.7252 & 0.4759 & 0.7220 & 0.6405 & 0.6310 & 0.5794 & 0.6170 & 0.6273 \\
Online-MetaRM & 0.7259 & 0.5196 & 0.7424 & 0.6364 & 0.6226 & 0.5846 & 0.6176 & \textbf{0.6356} \\
Only-MetaRM & 0.6970 & 0.3562 & 0.7109 & 0.6884 & 0.6214 & 0.5807 & 0.6456 & 0.6143 \\
\bottomrule
\end{tabular}
}
\end{table}

\section{Understanding Reward Modeling from Theoretical Perspective: From Statistical Correlation to Causal Alignment}
\label{appendix:theory}

\subsection{Notation and Idealized Reward Modeling}
\label{appendix:theory_notation}

We consider the classical pairwise preference reward task: given an input prompt $x$, the model generates two candidate responses $y_A, y_B \in \mathcal{Y}$, and human annotators provide a preference label $l \in \{A, B\}$.

For any triplet $(x, y_A, y_B)$, we assume there exists an ideal \textbf{evaluation dimension vector} $\mathbf{z} = [z_1, z_2, \dots, z_K]^\top$ and \textbf{weight vector} $\mathbf{w} = [w_1, w_2, \dots, w_K]^\top$, where each dimension $z_k$ characterizes an interpretable quality attribute (e.g., factual accuracy, logical coherence, linguistic politeness, response length, etc).
Weight $w_k \geq 0$ captures the importance of that dimension in the current comparison scenario.
The combination of dimensions and weights constitutes what we refer to as the reward causality.
For generative reward models, reward causality is sometimes contained in the critique rather than direct numerical computation.

\textbf{Note}: Both the dimension and weight vector $\mathbf{w}$ are context-dependent—only after given a specific $(x, y_A, y_B)$ can we infer which dimensions actually participate in human judgment and their relative importance. Ideally, the weights of key dimensions should approach 1, while irrelevant dimensions should have weights approaching 0. For example: in knowledge-intensive QA tasks, the weight of "factual accuracy" far exceeds "linguistic politeness"—a factually incorrect response, no matter how politely expressed, typically cannot prevail over a factually accurate but plain-toned response.

\subsection{Summary of Reward Modeling Mechanisms}

Since the emergence of traditional BT reward modeling, numerous improvement schemes have emerged to address the limitations of reward modeling. Their common approach is to enhance model performance through causal alignment.

This article formalizes "causality" as a combined representation of evaluation dimensions and weights, and systematically reviews the following schemes shown in Table~\ref{tab:summary_reward_modeling}. 

\begin{table}[h]
\centering
\small
\caption{Summary of Different Reward Modeling Mechanisms}
\label{tab:summary_reward_modeling}
\resizebox{1.0\textwidth}{!}{
\begin{tabular}{
    >{\raggedright\arraybackslash}m{2.6cm}
    >{\raggedright\arraybackslash}m{3cm}
    >{\raggedright\arraybackslash}m{8.1cm}
    >{\raggedright\arraybackslash}m{3.6cm}
}
\toprule
\multicolumn{1}{c}{\textbf{Method}} &
\multicolumn{1}{c}{\textbf{Modeling Approach}} &
\multicolumn{1}{c}{\textbf{Modeling Formula}} &
\multicolumn{1}{c}{\textbf{Learning Objective}} \\
\midrule
Bradley-Terry Reward Modeling & 
Compress high-dimensional evaluation space into a single scalar score & 
$P_\phi(y_A \succ y_B | x) = \sigma\left(r_\phi(x, y_A) - r_\phi(x, y_B)\right)$ & 
Learn scalar utility function through minimizing negative log-likelihood loss, such that $r_\phi(x, y) \approx \mathbf{w}^\top \mathbf{z}$ \\
\midrule
Conditional BT Reward Modeling & 
Explicitly model causality as part of the input & 
$P_\phi(y_A \succ y_B | x, \mathbf{d}, \mathbf{w}) = \sigma\left(\mathbf{w}^\top r_\phi(x, y_A | \mathbf{d}) - \mathbf{w}^\top r_\phi(x, y_B | \mathbf{d})\right)$ & 
Learn decomposed rewards conditioned on dimensions, based on predefined dimension description vector $\mathbf{d}$ and weight vector $\mathbf{w}$ \\
\midrule
Generative Reward Modeling with Outcome Supervision & 
Explicitly model causality as part of the output & 
$P_\theta(c, l | x, y_A, y_B) = P_\theta(c | x, y_A, y_B) \cdot \underbrace{P_\theta(l | x, y_A, y_B, c)}_{\text{supervised}}$ & 
Generate critique $c$ and preference label $l$, using only preference label correctness as supervision signal, expecting causal capability through result correctness constraint \\
\midrule
Generative Reward Modeling with Process Supervision & 
Explicitly supervise the causality in the output & 
$P_\theta(c, l | x, y_A, y_B) = \underbrace{P_\theta(c | x, y_A, y_B)}_{\text{supervised}} \cdot \underbrace{P_\theta(l | x, y_A, y_B, c)}_{\text{supervised}}$ & 
Directly optimize causal alignment for explicit critique supervision \\
\bottomrule
\end{tabular}
}
\end{table}

\subsection{Bradley-Terry Reward Modeling}

\textbf{Modeling Mechanism}: Traditional reward models assume a scalar utility function $r_\phi(x, y)$. The core idea is to compress the high-dimensional evaluation space $\mathbf{z}$ into a single scalar score. The preference probability follows the Bradley-Terry (BT) model:

\begin{equation}
P_\phi(y_A \succ y_B | x) = \sigma\left(r_\phi(x, y_A) - r_\phi(x, y_B)\right)
\end{equation}

The model is trained by minimizing the negative log-likelihood loss $\mathcal{L}_{\text{BT}}$. Ideally, the scalar reward should approximate the inner product of dimension features and true weights:

\begin{equation}
r_\phi(x, y) \approx \mathbf{w}^\top \mathbf{z}
\end{equation}

\textbf{Failure Modes}: The BT model is fundamentally a statistical learning approach that estimates preference patterns from empirical data distributions. While effective at the dataset level, this statistical nature introduces two primary sources of bias that compromise its discriminative power on individual samples:

\textbf{1. Sample-Level Causal Ambiguity}: The BT model suffers from causal sparsity of supervision: for a sample $(x, y_A, y_B, l)$, the training objective only models the preference label $l$ without capturing which dimensions determined the preference and their relative importance~\citep{ArmoRM,INF-ORM-Llama3.1-70B}. To compensate, the model relies on statistical aggregation extracted from patterns of similar samples in the training data. Learning from aggregated statistics rather than sample-specific criteria leads to inaccurate causal estimation at the individual sample level. 

\textbf{2. Dataset-Level Causal Ambiguity}: Beyond similar samples, global dataset patterns also introduce bias~\citep{liu2024rrm}. Since the BT model uses global empirical risk minimization for optimization, if a large proportion of "better" responses in the dataset share features like "high detail" or "polite tone," the model may increase their weights globally, even when irrelevant to specific samples. 

\textbf{3. Lack of Reasoning Capability}: Without reasoning capability, the BT model tends to capture surface-level correlations with label $l$, failing to model dimensions requiring deep reasoning~\citep{rmr1,rrm,liu2025inference}. For example, in math tasks, it struggles to learn "answer correctness" (which needs symbolic verification) and instead relies on proxy features like "solution detail."

In summary, the traditional BT model introduces causal ambiguity (or spurious correlations) for individual samples and lacks reasoning capability. It is only suitable for dimensions that are globally important and surface-observable (e.g., style, politeness, detail). 

\subsection{Conditional BT Reward Modeling}

\textbf{Modeling Mechanism}: To alleviate the causal ambiguity of BT reward modeling, recent work~\cite{wang2025rlbff} proposes explicitly modeling evaluation dimensions into the reward model. Although existing schemes mostly focus on single-dimension scenarios (implicitly assuming the weight of this dimension is 1), they can naturally extend to more general settings:

\begin{equation}
P_\phi(y_A \succ y_B | x, \mathbf{d}, \mathbf{w}) = \sigma\left(\mathbf{w}^\top r_\phi(x, y_A | \mathbf{d}) - \mathbf{w}^\top r_\phi(x, y_B | \mathbf{d})\right)
\end{equation}

where $\mathbf{d} = [d_1, d_2, \dots, d_K]$ is a predefined dimension description vector, $r_\phi(x, y | \mathbf{d}) \in \mathbb{R}^K$ is the decomposed reward conditioned on dimensions, and $\mathbf{w}$ is a manually specified weight vector.

\textbf{Failure Modes}: This mechanism directly addresses causal ambiguity by pre-specifying causal auxiliary evaluation. Several works~\cite{gu2026sparse,wang2025rlbff,yu2025rewardanything,wang2025scan} leverage LLM's reasoning capability for conditional generative reward modeling. Note that while these works address the lack of reasoning capability in BT models, all conditional reward modeling mechanisms face a significant limitation.

Specifically, since the dimension set $\mathbf{d}$ must be predefined, this method is only effective when key dimensions are highly determined. When facing (1) open-domain tasks with high openness such as open-domain dialogue and role-playing, and (2) tasks requiring complex and diverse process rewards such as multi-turn task planning and long-horizon reasoning, evaluation dimensions and weights are typically difficult to pre-specify, or even if pre-specified, are difficult to maintain consistency across all samples. Conditional reward modeling fails in these scenarios.

In summary, pre-specifying evaluation dimensions and weights endows BT models with the ability to incorporate causal reasoning for reward assignment. Compared to traditional BT models, they are more resistant to spurious correlations, but their applicability is narrow due to the need to pre-specify dimensions and weights. They are more suitable for tasks with clear rules, such as "syntax correctness" and "query efficiency" in SQL generation, or "citation reliability" and "argumentation completeness" in fact-checking.

\subsection{Generative Reward Modeling with Outcome Supervision}

\textbf{Modeling Mechanism}: Generative reward modeling changes the paradigm of reward modeling. Unlike conditional reward modeling which models causality in the input part, it models causality in the output part. Given input $(x, y_A, y_B)$, the model is required to generate a text sequence containing "critique" and "outcome". Some methods require the model to explicitly output evaluation dimensions and weights in the critique before giving the outcome~\citep{liu2025inference}, while others implicitly include this process in the critique without explicitly outputting related content~\citep{wang2025helpsteer3,rmr1,rrm,huang2025think,xu2025j4r}. 

\begin{equation}
P_\theta(c, l | x, y_A, y_B) = P_\theta(c | x, y_A, y_B) \cdot \underbrace{P_\theta(l | x, y_A, y_B, c)}_{\text{supervised}}
\end{equation}

where $c$ is a natural language analysis of the superiority of the two responses (i.e., critique), which explicitly points out the evaluation dimensions $\mathbf{z}$ and local weights $\mathbf{w}$ in the semantic space. $l \in \{A, B\}$ is the final preference label derived from the critique. Training typically uses correctness of the preference label as a supervision signal to optimize the probability of the model predicting the correct label $l$.

\textbf{Failure Modes}: Compared to conditional reward modeling, GRM has the following advantages: \textbf{(1) Automatic Causal Discovery}: By generating critique $c$, the model is forced to explicitly identify key dimensions of the current sample (e.g., code logic is more important than comments for this problem), achieving identification of key dimensions and dynamic weight allocation. Additionally, it can use test-time scaling to find better dimensions and weights. \textbf{(2) Resistance to Global Ambiguity}: Since GRM requires explicit causal analysis and learning for each sample, it is harder to be affected by global statistical biases.

Despite the great potential of GRM, our work identifies a critical misconception in existing approaches: they actually expect the model to learn good causal capability through outcome correctness. However, we find this introduces significant noise to training as discussed above.

\subsection{Generative Reward Modeling with Process Supervision}

To directly align reward modeling with causality, we introduce process supervision, i.e., supervision of critique. Our paradigm requires GRM to optimize the correctness of critiques. The modeling approach remains unchanged, but the supervision signal changes:

\begin{equation}
P_\theta(c, l | x, y_A, y_B) = \underbrace{P_\theta(c | x, y_A, y_B)}_{\text{supervised}} \cdot \underbrace{P_\theta(l | x, y_A, y_B, c)}_{\text{supervised}}
\end{equation}

Important evidence supporting this argument comes from the difference in critique word frequency distributions of models trained with the two supervision methods (see Section~\ref{sec:ablation_on_critique_evolution} and Table~\ref{tab:characteristic_terms} for details). 

\subsection{Theory-Practice Gap}

Although our theoretical framework indicates that we should conduct reward modeling by supervising critique, there are several challenges in practice: 

\textbf{Cost of High-Quality Human Critiques}: Currently, our approach to critique supervision relies on high-quality human critiques; however, collecting such data is both difficult and resource-intensive. 

\textbf{Difficulty in Obtaining Accurate Supervision}: We use the similarity between GRM-generated and human critiques as the supervision metric, achieving 85.71\% accuracy. This imperfect signal leads to: (1) suboptimal reward model performance, and (2) training instability. Fortunately, we have addressed this issue by introducing outcome regularization, which enables stable long-term training and ultimately achieves optimal performance. 

Despite these theory-practice gaps, we believe our work points to a promising direction for future reward modeling, i.e., through substantial engineering efforts to obtain higher-quality human critiques, thereby obtaining high-quality critique supervision signals and ultimately more accurate reward models. Additionally, using human critique supervision may not be necessary, but this requires further research.

\newpage
\section{Prompt}
\label{appendix:prompt}

\begin{figure*}[!h]
\centering
\begin{promptbox}[Prompt: Template of Generative Reward Models]{blue!60!white}
\small
Please act as an impartial judge and evaluate the quality of the responses provided by two AI Chatbots to the Client question displayed below.

\medskip
[Client Question]

\{conv\_his\}

\medskip
[The Start of Chatbot A's Response]

\{response\_A\}

[The End of Chatbot A's Response]

\medskip
[The Start of Chatbot B's Response]

\{response\_B\}

[The End of Chatbot B's Response]

\medskip
Output your final verdict by strictly following this format:

\medskip
\texttt{<critics>}

[Provide a brief summary of your reasoning for the choice]

\texttt{</critics>}

\medskip
\texttt{<choice>}

[[A]]

\texttt{</choice>}

\medskip
Note: Use [[A]] if A is better, or [[B]] if B is better.
\end{promptbox}
\caption{Prompt: Template of Generative Reward Models.}
\label{fig:prompt_main_prompt_grm}
\end{figure*}

\begin{figure*}[!h]
\centering
\begin{promptbox}[Prompt: Edit-only Refinement Instruction based on Critiques]{purple!60!white}
\small
\textbf{[SYSTEM RULE: EDIT-ONLY MODE ENGAGED]}

You are a text-processing bot. You are FORBIDDEN from answering the question. Your only function is to apply edits.

\medskip
\textbf{PRIMARY DIRECTIVE: Follow the \texttt{<critique>}. Nothing else.}

\begin{enumerate}
\item \textbf{THE EXCEPTION RULE:} IF the \texttt{<critique>} states something is factually or mathematically wrong, you are authorized to fix \textbf{ONLY THAT SINGLE PIECE OF INFORMATION.} Do not explain. Do not add context. Just replace the wrong data with the correct data.

\item \textbf{THE DEFAULT RULE:} For everything else, if the \texttt{<critique>} does not explicitly order a change, you MUST NOT change it. Do not fix other errors. Do not improve style. Do not add information.

\item \textbf{THE RE-CHECK RULE:} Before you respond, you must check again whether the \texttt{<critique>} contain the content you modify.
\end{enumerate}

Failure to follow these rules means you fail the task.

\rule{\linewidth}{0.4pt}

[Client Question]

\{conv\_his\}

\medskip
\texttt{[The Start of Chatbot A's Response]}

\{response\_A\}

[The End of Chatbot A's Response]

\medskip
[The Start of Chatbot B's Response]

\{response\_B\}

[The End of Chatbot B's Response]

\medskip
[The Start of Critique]

\{critique\}

[The End of Critique]
\end{promptbox}
\caption{Prompt: Edit-only Refinement Instruction based on Critiques.}
\label{fig:prompt_edit_model}
\end{figure*}

\begin{figure*}[!h]
\centering
\begin{promptbox}[Prompt: Calculation of Similarity w/ Core Human Critiques]{orange!60!white}
\small
I will provide you with a generated evaluation content and a reference evaluation content. Your task is to analyze the similarity between the <Generated Evaluation Content> and the <Reference Evaluation Content> by calculating F1 scores based on their key arguments.

\textbf{Core Principle:} Focus exclusively on ``Key Arguments'' - decisive reasons that are powerful enough to justify the final choice on their own. Identify these core justifications, not minor points.

\medskip
\#\# Part 1: First check

First check if the generated critique repeats the same point across multiple times. If yes, directly output without conducting part 2:

\begin{verbatim}
<thinking>
    Put here how the generated critique repeats points.
</thinking>

<scores>
    <critique_f1>0</critique_f1>
    <critique_precision>0</critique_precision>
    <critique_recall>0</critique_recall>
</scores>
\end{verbatim}

\medskip
\#\# Part 2: Steps for F1 Score Calculation

\begin{enumerate}
\item \textbf{Count Reference Key Arguments (N\_ref):}
    \begin{itemize}
    \item Check if the reference identifies a \textbf{fatal error} (critical factual error, harmful statement, or fundamental misunderstanding).
        \begin{itemize}
        \item \textbf{If yes}: Only this fatal error counts. Set \textbf{N\_ref = 1}.
        \item \textbf{If no}: Count all unique Key Arguments (decisive reasons that could justify the choice by themselves). Set \textbf{N\_ref} to this count.
        \end{itemize}
    \end{itemize}

\item \textbf{Count Generated Key Arguments (N\_gen):}
    \begin{itemize}
    \item Identify all unique Key Arguments in the generated evaluation.
    \item Set \textbf{N\_gen} to this count.
    \end{itemize}

\item \textbf{Count True Positives (TP):}
    \begin{itemize}
    \item Initialize \texttt{TP = 0}.
    \item For each reference key argument, search for a match in generated key arguments.
    \item \textbf{Matching Rule:} Both \textbf{semantic meaning} and \textbf{stance} (which response and positive/negative) must align.
        \begin{itemize}
        \item Example: ``Response A is more detailed'' only matches with similar praise of Response A, not Response B.
        \item For fatal errors: Generated must identify the same error in the same response.
        \end{itemize}
    \item Each generated argument can only match once.
    \item Increment \texttt{TP} by 1 for each valid match.
    \end{itemize}

\item \textbf{Calculate Scores:}
    \begin{itemize}
    \item \textbf{Precision\_critique:} \texttt{TP / N\_gen} (0 if N\_gen = 0)
    \item \textbf{Recall\_critique:} \texttt{TP / N\_ref} (0 if N\_ref = 0)
    \item \textbf{CritiqueScore:} \texttt{2 * (Precision * Recall) / (Precision + Recall)} (0 if sum = 0)
    \end{itemize}
\end{enumerate}

\textbf{Output Format (rounded to 4 decimal places):}

\begin{verbatim}
<thinking>
    Put the thinking process here.
</thinking>

<scores>
    <critique_f1>CritiqueScore</critique_f1>
    <critique_precision>Precision_critique</critique_precision>
    <critique_recall>Recall_critique</critique_recall>
</scores>
\end{verbatim}

\texttt{<Generated Evaluation Content>}

\{critiques\}

\texttt{</Generated Evaluation Content>}

\medskip
\texttt{<Reference Evaluation Content>}

\{reference\_critiques\}

\texttt{</Reference Evaluation Content>}
\end{promptbox}
\caption{Prompt: Calculation of Similarity w/ Core Human Critiques.}
\label{fig:prompt_similarity_core_hc}
\end{figure*}

\begin{figure*}[!h]
\centering
\begin{promptbox}[Prompt: Calculation of Similarity w/ All Human Critiques]{orange!60!white}
\small
I will provide you with a generated evaluation content and a reference evaluation content. Your task is to analyze the similarity between the \texttt{<Generated Evaluation Content>} and the \texttt{<Reference Evaluation Content>} by calculating F1 scores based on their all arguments.

\medskip
\#\# Part 1: First check

First check if the generated critique repeats the same point across multiple times. If yes, directly output without conducting part 2:

\begin{verbatim}
<thinking>
    Put here how the generated critique repeats points.
</thinking>

<scores>
    <critique_f1>0</critique_f1>
    <critique_precision>0</critique_precision>
    <critique_recall>0</critique_recall>
</scores>
\end{verbatim}

\medskip
\#\# Part 2: Steps for F1 Score Calculation

\begin{enumerate}
\item \textbf{Count Reference All Arguments (N\_ref):}
    \begin{itemize}
    \item Check if the reference identifies a \textbf{fatal error} (critical factual error, harmful statement, or fundamental misunderstanding).
        \begin{itemize}
        \item \textbf{If yes}: Only this fatal error counts. Set \textbf{N\_ref = 1}.
        \item \textbf{If no}: Count all unique Arguments (decisive reasons that could justify the choice by themselves). Set \textbf{N\_ref} to this count.
        \end{itemize}
    \end{itemize}

\item \textbf{Count Generated All Arguments (N\_gen):}
    \begin{itemize}
    \item Identify all unique Arguments in the generated evaluation.
    \item Set \textbf{N\_gen} to this count.
    \end{itemize}

\item \textbf{Count True Positives (TP):}
    \begin{itemize}
    \item Initialize \texttt{TP = 0}.
    \item For each reference argument, search for a match in generated arguments.
    \item \textbf{Matching Rule:} Both \textbf{semantic meaning} and \textbf{stance} (which response and positive/negative) must align.
        \begin{itemize}
        \item Example: ``Response A is more detailed'' only matches with similar praise of Response A, not Response B.
        \item For fatal errors: Generated must identify the same error in the same response.
        \end{itemize}
    \item Each generated argument can only match once.
    \item Increment \texttt{TP} by 1 for each valid match.
    \end{itemize}

\item \textbf{Calculate Scores:}
    \begin{itemize}
    \item \textbf{Precision\_critique:} \texttt{TP / N\_gen} (0 if N\_gen = 0)
    \item \textbf{Recall\_critique:} \texttt{TP / N\_ref} (0 if N\_ref = 0)
    \item \textbf{CritiqueScore:} \texttt{2 * (Precision * Recall) / (Precision + Recall)} (0 if sum = 0)
    \end{itemize}
\end{enumerate}

\textbf{Output Format (rounded to 4 decimal places):}

\begin{verbatim}
<thinking>
    Put the thinking process here.
</thinking>

<scores>
    <critique_f1>CritiqueScore</critique_f1>
    <critique_precision>Precision_critique</critique_precision>
    <critique_recall>Recall_critique</critique_recall>
</scores>
\end{verbatim}

\texttt{<Generated Evaluation Content>}

\{critiques\}

\texttt{</Generated Evaluation Content>}

\medskip
\texttt{<Reference Evaluation Content>}

\{reference\_critiques\}

\texttt{</Reference Evaluation Content>}
\end{promptbox}
\caption{Prompt: Calculation of Similarity w/ All Human Critiques.}
\label{fig:prompt_similarity_full_hc}
\end{figure*}

\begin{figure*}[!h]
\centering
\begin{promptbox}[Prompt: LLM-as-a-Meta-Judge]{orange!60!white}
\small
You are an expert evaluator tasked with assessing the quality of critiques comparing two responses.

You will be given:
\begin{enumerate}
\item A conversation history
\item Response A
\item Response B
\item One or more critiques comparing Response A and Response B
\end{enumerate}

Your task is to evaluate whether the critique(s) are accurate and correct based on the actual content of the responses. Assign a score between 0 and 1, where higher scores indicate more accurate critiques.

You must provide your response in the following XML format:

\begin{verbatim}
<thinking>
    Put your detailed analysis here. Examine the critique 
    against the actual responses and explain your reasoning 
    for the score.
</thinking>

<scores>
    <critique_f1>Score</critique_f1>
    <critique_precision>Score</critique_precision>
    <critique_recall>Score</critique_recall>
</scores>
\end{verbatim}

Note: Assign the same score to all three metrics (critique\_f1, critique\_precision, and critique\_recall).

\medskip
\texttt{<Conversation History>}

\{conv\_his\}

\texttt{</Conversation History>}

\medskip
\texttt{<Response A>}

\{response\_A\}

\texttt{</Response A>}

\medskip
\texttt{<Response B>}

\{response\_B\}

\texttt{</Response B>}

\medskip
\texttt{<Critiques>}

\{critiques\}

\texttt{</Critiques>}
\end{promptbox}
\caption{Prompt: LLM-as-a-Meta-Judge.}
\label{fig:prompt_llm_as_a_meta_judge}
\end{figure*}